\documentclass[lettersize,journal]{IEEEtran}
\usepackage{amsmath,amsfonts}
\usepackage{algorithmic}
\usepackage{algorithm}
\usepackage{array}
\usepackage[caption=false,font=normalsize,labelfont=sf,textfont=sf]{subfig}
\usepackage{textcomp}
\usepackage{stfloats}
\usepackage{url}
\usepackage{verbatim}
\usepackage{graphicx}
\usepackage{cite}
\usepackage{amsfonts}
\usepackage{xcolor}
\usepackage{xspace} %

\def\onedot{\ifx\@let@token.\else.\null\fi\xspace} 
\def\eg{\emph{e.g}\onedot}

\newcommand{\ie}[0]{\emph{i.e.}}

\makeatother 

\newcommand{\ourmethodCC}[0]{{MAIL}~}

\usepackage{booktabs}  

\usepackage{amsmath}

\usepackage{colortbl}
\definecolor{bgreen}{RGB}{0,170,0}
\definecolor{bred}{RGB}{220,0,0}
\definecolor{mydarkblue}{RGB}{0,0,150}
\definecolor{Gray}{gray}{0.92}
\definecolor{Grey}{gray}{0.85}
\definecolor{darkGrey}{gray}{0.66}
\definecolor{lightPurple}{rgb}{0.88,0.88,0.96} 
\definecolor{Purple}{rgb}{0.6,0.6,0.96}
\definecolor{y}{RGB}{255, 250, 205}
\definecolor{p}{RGB}{245, 234, 240}
\definecolor{b}{RGB}{224, 255, 255}
\definecolor{mytableblue}{rgb}{0.83, 0.90, 0.94}
\definecolor{mytablegreen}{rgb}{0.90, 0.97, 0.87}

\usepackage{hyperref} 
\hypersetup{  
    colorlinks=true,
    linkcolor=bred,
    citecolor=mydarkblue,
    filecolor=bred,
    urlcolor=mydarkblue
}

\usepackage{wrapfig}

\usepackage{algorithm}
\usepackage{makecell}
\usepackage{float}  

\definecolor{PineGreen}{HTML}{008B00}
\definecolor{BrickRed}{HTML}{B22222}

\usepackage[textsize=tiny]{todonotes}
\usepackage{bbm} 
\usepackage{graphicx} 

\usepackage{multirow} 

\usepackage{pifont}  

\usepackage{svg} 
\usepackage{listings}   

\begin{document}

\title{MAIL++: Multi-Modal Bi-directional Agent Layer for Vision-Language Models}

\author{Kaixiang Chen, Pengfei Fang$^*$, and Hui Xue$^*$,~\IEEEmembership{Member,~IEEE}

\thanks{

K. Chen, P. Fang and H. Xue are with the School of Computer Science and Engineering, Southeast University, Nanjing 210096, China and Key Laboratory of New Generation Artificial Intelligence Technology and Its Interdisciplinary Applications (Southeast University), Ministry of Education, China (E-mail: kxchen, fangpengfei, hxue@seu.edu.cn).

$^*$Co-corresponding author}
}



\maketitle


\begin{abstract}

Adapting large   vision–language models (VLMs) such as CLIP to downstream tasks remains non-trivial, as full-model fine-tuning is computationally prohibitive and prone to severe overfitting in low-data regimes. Parameter-efficient fine-tuning (PEFT) methods alleviate these challenges by introducing lightweight prompt- or adapter-based modules, yielding superior performance. Among them, cross-modal coupling strategies have demonstrated particular effectiveness by explicitly enhancing inter-modal interactions. 
However, existing coupling mechanisms are predominantly rely on external auxiliary modules, resulting in indirect and coarse-grained interactions that remain structurally decoupled from the original VLM, which fundamentally limit their representational expressiveness.
In this paper, we propose Multi-Modal Interactive Agent Layer (MAIL), a new PEFT paradigm that embeds cross-modal coupling directly \textit{within} the VLM's intrinsic computation modules.
Specifically, MAIL freezes the backbone and inserts lightweight agent layers after core computational modules (e.g., LayerNorm) to approximate the parameter updates that full fine-tuning would induce. To couple the visual and textual streams at the level of these computation modules, we introduce a bottleneck-based text-to-image bridge that couples the optimization of paired agent layers across modalities, effectively coordinating the adaptation of corresponding computation modules.
Building on MAIL, we further present MAIL++, which enables bi-directional cross-modal exchange via a meta agent layer together with a meta-text bridge and a meta-image bridge. At inference time, all agent layers are re-parameterized into the frozen backbone, preserving the original computational efficiency. Extensive experiments on few-shot image classification and few-shot universal cross-domain retrieval benchmarks show that MAIL and MAIL++ consistently outperforms state-of-the-art PEFT methods.

\end{abstract}

\begin{IEEEkeywords}
Vision-Language Models, Parameter-Efficient Fine-Tuning, Cross-Modal Coupling.
\end{IEEEkeywords}


\section{Introduction}


\IEEEPARstart{F}{oundation} vision-language models (VLMs) have emerged as a paradigm-shifting approach in visual representation learning, marking a critical step toward more general-purpose learning systems~\cite{foundation}. Among them, CLIP~\cite{CLIP} serves as a representative VLM, featuring an image encoder and a text encoder trained with a contrastive objective to align images and texts at scale using millions of paired samples. In contrast to conventional ImageNet-era architectures~\cite{ResNet, ViT, he}, which typically rely on task-specific fine-tuning with large training dataset for downstream adaptation, VLMs like CLIP exhibit strong zero-shot generalization capability. This property has catalyzed their deployment across a broad spectrum of downstream applications, including medical image analysis~\cite{medical-1, medical-2, medical-3}, cross-modal and cross-domain retrieval~\cite{ProS, DePro, DCLIP}, and human activity recognition~\cite{Activity-analysis-1, Activity-analysis-2, Activity-analysis-3}.
While it is also reasonable to further fine-tune VLMs on downstream tasks using task-specific data for better performance, fine-tuning such large-scale pretrained models remains non-trivial, owing to two key challenges: \textit{(i)} the prohibitive computational cost associated with full-model optimization, and \textit{(ii)} the pronounced risk of overfitting, which is particularly severe in data-scarce regimes commonly encountered in practice.

\begin{figure*}[t]
\begin{center}
\centerline{\includegraphics[width=0.95\textwidth]{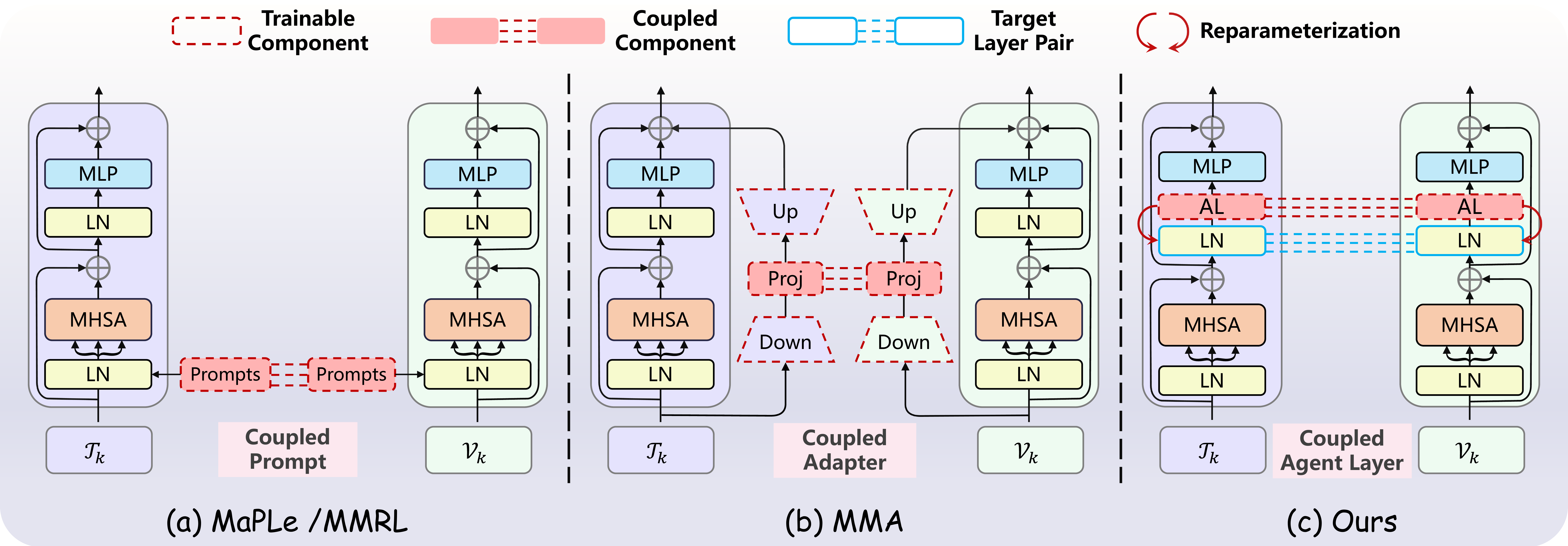}}
\caption{Modality-coupled methods in fine-tuning VLMs:
(a) MaPLe and MMRL achieve cross-modal coupling by establishing interconnections between the prompts.
(b) MMA designs a unified feature-projection layer within the adapter that is shared by both modalities.
(c) In contrast, MAIL achieves cross-modal coupling while preserving inference efficiency by introducing linked trainable agent layers (AL) to align internal parameter updates of the intrinsic computational modules between the visual and textual streams within the backbone. }
\vspace{-0.5\baselineskip}
\label{fig: Multi-Modal Designs}
\end{center}
\end{figure*}

These challenges have spurred the development of parameter-efficient fine-tuning (PEFT) methods for few-shot VLM adaptation~\cite{PEFT}. A particularly active line of research is prompt-based tuning~\cite{CoOp, CoCoOp, KgCoOp, PromptSRC, MaPLe, MMRL}. CoOp~\cite{CoOp} pioneers continuous prompt learning~\cite{prompt-tuning} by representing prompt tokens as learnable vectors and injecting them into the text encoder, yet it often overfits to seen classes in low-data regimes. To improve generalization, CoCoOp~\cite{CoCoOp} introduces a meta-network that produces instance-conditioned prompts, enhancing transfer to unseen classes. Subsequent methods, such as KgCoOp~\cite{KgCoOp} and PromptSRC~\cite{PromptSRC}, further mitigate overfitting by incorporating knowledge-guided regularization. More recently, MaPLe~\cite{MaPLe} and MMRL~\cite{MMRL} extend prompt tuning to both the image and text encoders and explicitly couple the learned prompts across modalities, yielding stronger and more consistent gains on both seen and unseen categories (\textbf{Fig.~\ref{fig: Multi-Modal Designs}-a}). Complementary to prompt tuning, adapter-based PEFT methods~\cite{Tip-adapter, Clip-adapter, MMA} inject lightweight modules to refine intermediate representations without modifying the input space. Notably, coupling adapters across modalities (\eg, MMA; \textbf{Fig.~\ref{fig: Multi-Modal Designs}-b}) consistently outperforms their modality-isolated counterparts, further substantiating the critical role of cross-modal interaction in effective VLM adaptation.

We categorize these approaches (\ie, MaPLe, MMRL and MMA) as modality-coupled methods (MCMs), and refer to their counterparts that adapt each modality independently as modality-independent methods (MIMs). Explicit cross-modal coupling endows MCMs with two primary advantages. \textbf{\textit{First}}, since the objective of vision–language training is to establish robust inter-modal alignment, explicitly modeling cross-modal interactions reinforces this alignment and promotes effective knowledge transfer to target domains. \textbf{\textit{Second}}, modality coupling serves as an implicit regularization mechanism. By enforcing coordinated updates across modalities, MCMs encourage the learning of coherent and mutually consistent representations, thereby mitigating overfitting and improving generalization. Despite these benefits, it is important to note that the coupling mechanisms employed in existing MCMs are predominantly realized through \textit{external auxiliary components}, such as input-level prompts or adapter layers, rather than being embedded in the intrinsic computational modules of the original VLM. While such externally mediated coupling strategies are lightweight and empirically effective, the corresponding coupling signals must propagate through multiple layers before influencing the deep semantic representations of the intrinsic computational modules. 
Consequently, the induced cross-modal interactions tend to be coarse-grained and indirect, remaining structurally
decoupled from the original VLM, 
which fundamentally limit the its capacity for fine-grained cross-modal alignment.

 To bridge this gap, we propose Multi-Modal Interactive Agent Layer (MAIL), a MCM designed to embed cross-modal coupling directly \textit{within} the VLM's intrinsic computation modules, as illustrated in \textbf{Fig.~\ref{fig: Multi-Modal Designs}-c}. 
 Concretely, for each selected pair of corresponding computation modules in the image and text encoders, MAIL designates them as a target module pair and freezes their parameters. For each target module, a lightweight agent layer is appended to functionally approximate the localized parameter updates of the target module that full fine-tuning would induce.
 Each agent layer consists of a \textit{scaling} component and a \textit{shifting} component~\cite{SSF}, which are jointly optimized to capture fine-grained adjustments of the target module. To couple the visual and textual streams at the level of computation modules, we introduce a bridge function that links the agent layers across modalities, effectively coordinating the adaptation
of corresponding computation module.
 

\begin{figure}[t]
\begin{center}
\centerline{\includegraphics[width=0.95\columnwidth]{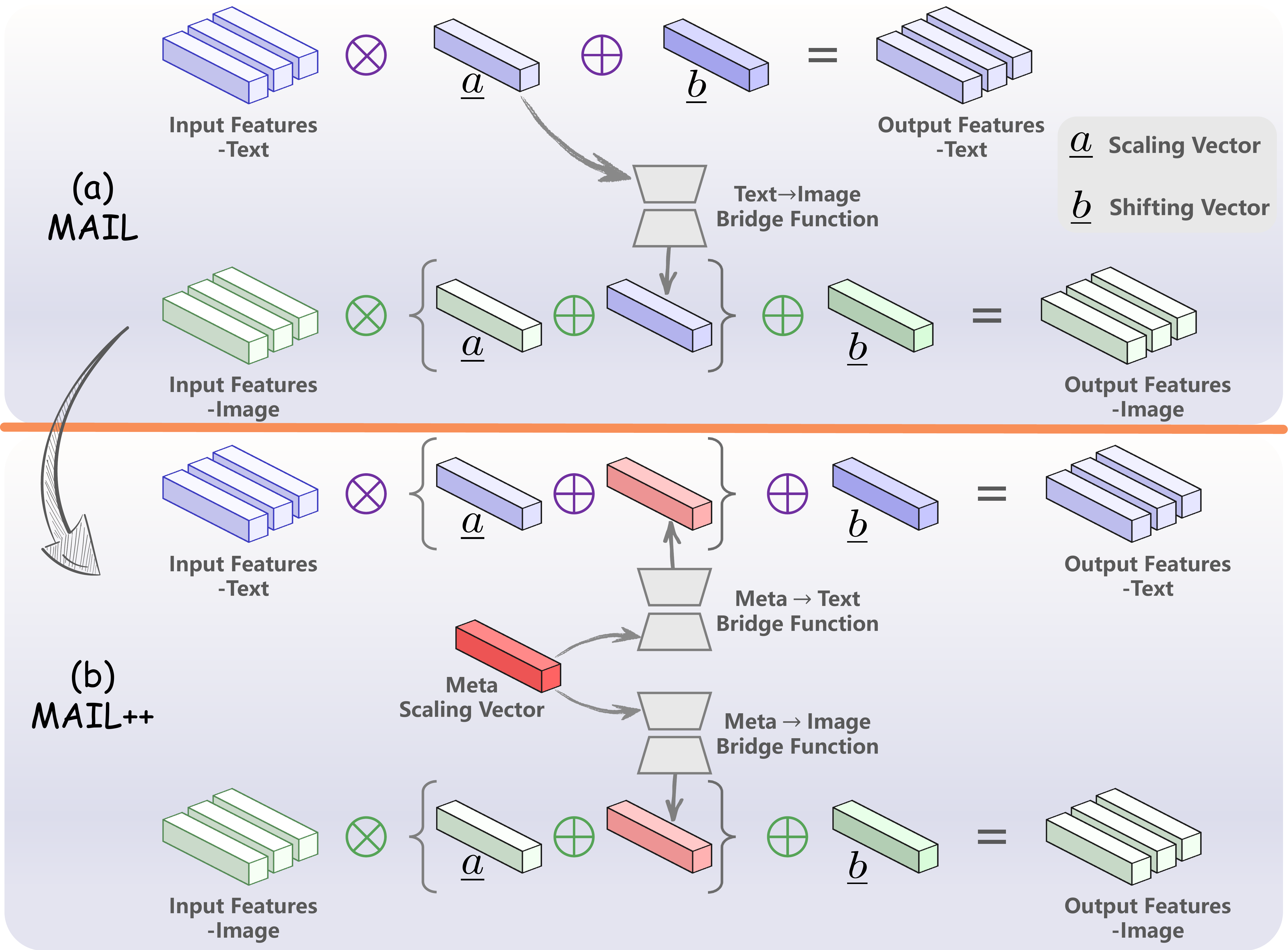}}
\caption{Illustration of the coupling mechanisms in the agent layer. (a) The scaling vector from the text-side agent layer propagates to the vision side via a text-to-image bridge function. (b) A meta scaling vector is incorporated into both sides via meta-to-image and meta-to-text bridge functions, establishing a bidirectional information flow.}
\vspace{-0.3\baselineskip}
\label{fig: mailvsmail++}
\end{center}
\end{figure}

In our previous work~\cite{MAIL}, the bridge function was instantiated as a unidirectional text-to-image information flow, as shown in \textbf{Fig.~\ref{fig: mailvsmail++}-a}. This manually imposed inductive bias restricted cross-modal information propagation to a single direction, causing the optimization process to be dominated by the text modality and resulting in an imbalance between the two modalities. To overcome this limitation, we further propose MAIL++, which establishes a bidirectional cross-modal interaction mechanism that enables mutual information exchange and joint optimization between the two encoders. As illustrated in \textbf{Fig.~\ref{fig: mailvsmail++}-b}, rather than symmetrically introducing an additional image-to-text bridge, MAIL++ incorporates a meta-scaling vector, together with a meta-text bridge and a meta-image bridge, to dynamically regulate the magnitude and direction of cross-modal information flow. This design removes the hand-crafted directional bias and allows the model to adaptively learn balanced and coherent representations across modalities.

During inference, the agent layers are seamlessly reparameterized into their corresponding frozen layers, ensuring that the overall computational complexity remains unchanged. To summarize, our main contributions are outlined as follows: 
\begin{enumerate}
    \item We introduce MAIL, a novel modality-coupling mechanism (MCM) that, for the first time, explicitly integrates such mechanisms within the internal modules of the backbone to directly synchronize parameter updates across image and text modalities.
    \item We further propose MAIL++, an enhanced version of MAIL, which facilitates a more balanced information flow and mutual optimization between the modalities.
    \item Extensive experiments on eleven few-shot classification datasets and three FS-UCDR benchmarks (first introduced in our previous work~\cite{MAIL})  demonstrate that MAIL++ achieves state-of-the-art performance while maintaining superior parameter efficiency and inference efficiency, \textit{under the strict constraint that no LLMs or distillation techniques from larger backbones are employed}.
\end{enumerate}


\section{Related Work}

\subsection{Vision-Language Models}  
Recent progress in vision–language models (VLMs) has substantially advanced unified representation learning across visual and textual modalities. CLIP~\cite{CLIP} established the foundation for large-scale contrastive pre-training, yielding strong zero-shot generalization across open-vocabulary tasks. ViLT~\cite{ViLT} introduced a streamlined fully Transformer-based architecture that eliminates convolutions for efficient cross-modal processing, whereas ALBEF~\cite{ALBEF} coupled contrastive objectives with a cross-modal fusion encoder to enhance retrieval and captioning performance.
VLMo~\cite{VLMO} unified unimodal and multimodal encoders through an optimal-transport-based mixture-of-experts design, while BLIP~\cite{BLIP} leveraged bootstrapped captioning signals to improve pre-training efficiency and sample coverage. The recent LLaVA family~\cite{LLAVA} further integrates visual encoders with large language models (LLMs), enabling instruction-following multimodal reasoning.


\subsection{Parameter Efficient Fine-Tuning for VLMs} 
Although current vision–language models (VLMs) yield highly transferable and semantically rich representations, effectively adapting them to downstream tasks without degrading their inherent representational priors remains an open challenge. To address this issue, parameter-efficient fine-tuning (PEFT)~\cite{2019nlpAdapter, LoRA, BitFit, SSF} has emerged as a principled and effective adaptation paradigm for large-scale VLMs. Existing PEFT approaches for VLMs can be broadly categorized into two representative families: prompt- and adapter-based methods.

\textit{\textbf{Prompt-based methods.}} 
These approaches introduce a small number of learnable prompt tokens during adaptation while keeping the backbone frozen. CoOp~\cite{CoOp} pioneers continuous prompt learning by optimizing soft prompt embeddings appended to the text encoder. Building upon this formulation, CoCoOp~\cite{CoCoOp} incorporates a meta-network to generate instance-adaptive prompts, thereby improving generalization to unseen categories. From a complementary perspective, PromptSRC~\cite{PromptSRC} and KgCoOp~\cite{KgCoOp} mitigate overfitting by enforcing knowledge-guided regularization, encouraging the learned prompts to align with handcrafted textual templates and enhancing robustness. More recent approaches, such as MaPLe~\cite{MaPLe} and MMRL~\cite{MMRL}, further extend this line by inserting prompts into both the visual and textual encoders and explicitly coupling them across modalities, demonstrating consistent performance gains across seen and unseen classes.


\textit{\textbf{Adapter-based methods.}} 
 In contrast, adapter-based approaches introduce lightweight trainable modules within the network while keeping the pretrained CLIP parameters frozen. These adapters take various forms, including bottleneck architectures~\cite{MMA, MMA++}, residual layers~\cite{Clip-adapter}, residual prototype branches~\cite{APE, TaskRes}, and feature-memory mechanisms~\cite{Tip-adapter, CaFo}. CLIP-Adapter~\cite{Clip-adapter} augments CLIP’s dual-encoder architecture with lightweight two-layer MLP adapters, enabling task-driven recalibration of intermediate representations under cross-entropy supervision. Tip-Adapter~\cite{Tip-adapter} and CaFo~\cite{CaFo} further reduce adaptation overhead by caching training features and performing similarity-based retrieval at inference time. Meanwhile, APE~\cite{APE} and TaskRes~\cite{TaskRes} preserve the pretrained classifier and introduce lightweight residual branches to achieve task-specific adaptation while maintaining CLIP’s inherent priors. 
Despite their effectiveness, the majority of adapter-based approaches adapt the visual and textual encoders in a modality-isolated manner, which inherently limits cross-modal alignment. To alleviate this limitation, MMA~\cite{MMA} introduces a shared projection space to couple adapters across modalities, enabling cross-modal gradient propagation and fostering more coherent vision–language representations.


\section{Preliminary}
\subsection{Revisiting CLIP}
CLIP~\cite{CLIP} is a pretrained vision-language  model (VLM) consisting of two encoders: a text encoder, denoted by $\mathcal{T}(\cdot)$, and an image encoder (ViT~\cite{ViT} as default), denoted by $\mathcal{V}(\cdot)$. 
Both encoders comprise $L$ transformer~\cite{attention} blocks, represented as  $\{\mathcal{T}_i \}_{i=1}^{L}$ and $\{\mathcal{V}_i \}_{i=1}^{L}$, respectively. For classification inference with $C$ classes, CLIP inserts all class names into a pre-defined text template, e.g., ``\texttt{a photo of a $<$category$>$}'', generating $C$ inputs $\{t_i\}_{i=1}^{C}$ for the text encoder $\mathcal{T}(\cdot)$. For a certain input $t_y$, its output $\mathcal{T}(t_y)$ is:
\begin{equation}
\begin{aligned}
    & \mathcal{W}_0 = {\rm TextEmbed} (t_y)  \\
    & \mathcal{W}_i = \mathcal{T}_i(\mathcal{W}_{i-1}), \quad i = 1,2,...,L  \\
    & \mathcal{T}(t_y) = {\rm TextProj} \circ {\rm LN}(w_L^{N_t}),
\end{aligned}
\end{equation}
where $\mathcal{W}_0\!\!=\!\![w_0^0,w_0^1,...,w_0^{N_t}]^{\top}\!\!\in \!\mathbb{R}^{N_t\times d_t}$ is the word embedding, with $N_t$ and $d_t$ indicate text embedding length and dimension, $\circ$
represents the composition of functions. $ {\rm TextProj}$ is a linear layer ($d_t \rightarrow d_t$). 
Similarly, for the image $I$, its representation $ \mathcal{V}(I)$ is calculated as:    
\begin{equation}
\begin{aligned}
    & \mathcal{P}_0 = {\rm PatchEmbed} (I)  \\
    & [c_i, \mathcal{P}_i] = \mathcal{V}_i([c_{i-1}, \mathcal{P}_{i-1}]), \quad i = 1,2,...,L  \\
    & \mathcal{V}(I) =  {\rm ImageProj} \circ {\rm LN} (c_L),
\end{aligned}
\end{equation}
where $\mathcal{P}_0\! \in \!  \mathbb{R}^{N_v\times d_v}$ is the image embedding, with $N_v$ and $d_v$ indicate embedding length and dimension, and $c_0 \! \in \! \mathbb{R}^{d_v} $ is the initial $\rm CLS$ Token. $ {\rm ImageProj}$ is also a linear layer ($d_v\! \rightarrow \! d_t$). With $\mathcal{V}(I)$ available, the text features of the text templates with class labels are matched using the formula 
    $  p( y | I) \! =\! \frac{ \exp(sim(\mathcal{T}(t_{ y}),  \mathcal{V}(I)))}{\sum_{i=1}^C \exp(sim(\mathcal{T}(t_i), \mathcal{V}(I))) }, $
 where $ y \! \in \! \{1,2,..., C\}$, and $sim(., .)$ refers to  cosine similarity.
\section{Multi-Modal Interactive Agent Layer}
\label{MAIL}

\begin{figure*}[t]
\begin{center}
\centerline{\includegraphics[width=0.85\textwidth]{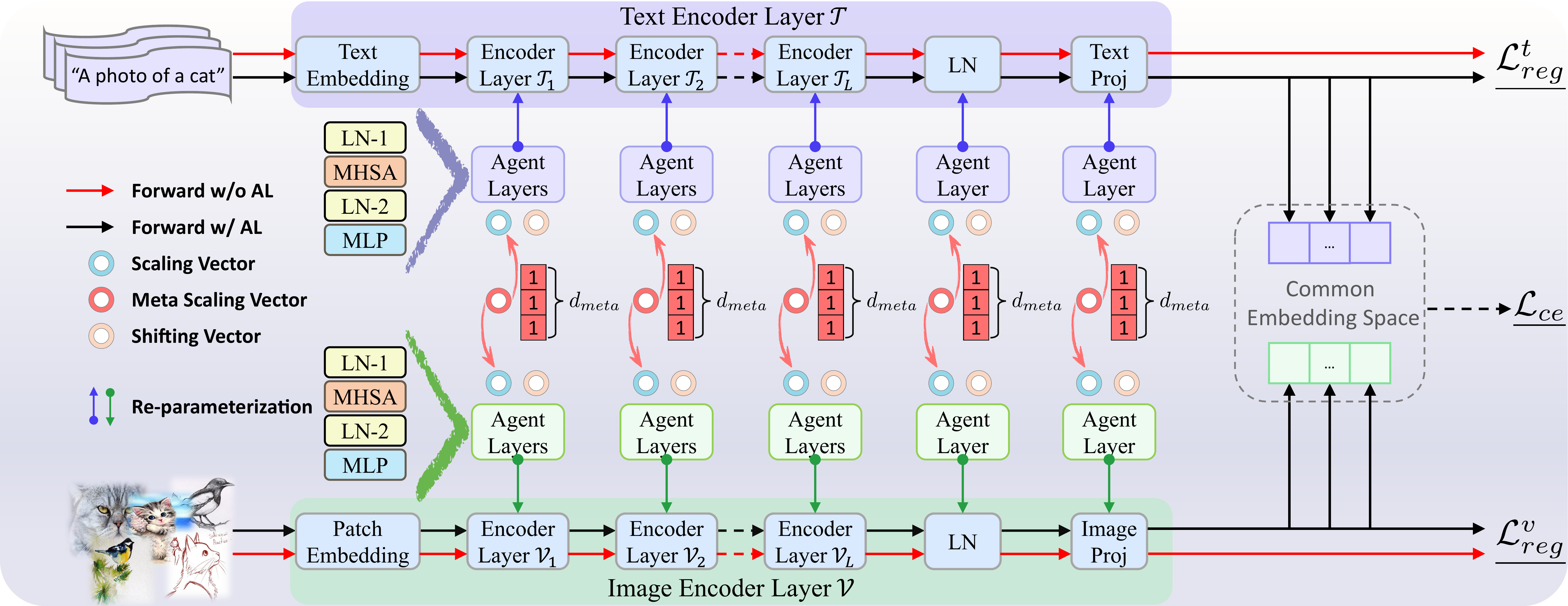}}
\vspace{-0.3\baselineskip}
\caption{The proposed Multi-Modal Interactive Agent Layer++ (MAIL++) for the transformer-based CLIP models. During training, we only fine-tune the agent layers, which are inserted into both encoders. The image agent layers interact with the text agent layers through a trainable bottleneck-based bridge function, fostering mutual synergy between the two modalities. }
\label{fig: MAIL}
\end{center}
\vspace{-1.0\baselineskip}
\end{figure*}

Modality-coupled methods (MCMs), such as MaPLe \cite{MaPLe} and MMA \cite{MMA}, enhance cross-modal alignment by strengthening the interactions between the image and text encoders during training, thereby improving downstream performance. However, the coupling mechanisms employed in existing MCMs are predominantly realized through external components, rather than being integrated into the intrinsic architecture. Consequently, the induced cross-modal interactions remain relatively coarse and indirect, which may constrain the model’s ability to capture fine-grained alignment.
To solve this problem, we propose the \textbf{M}ulti-Mod\textbf{a}l \textbf{I}nteractive Agent \textbf{L}ayer (MAIL), a lightweight MCM that aligns parameter updates of the intrinsic architecture within the backbone across modalities while preserving the inference efficiency. MAIL incorporates agent layers to capture localized parameter updates, while the bridge functions further refine and align these updates across encoders. Based on MAIL, we further propose MAIL++, as illustrated in \textbf{Fig.~\ref{fig: MAIL}}. MAIL++ establishes a bidirectional information flow between the two encoders,
enabling mutual interaction and joint optimization.

\vspace{-0.5\baselineskip}
\subsection{Agent Layer}
The agent layer (AL) is designed to approximately capture the updates of specific computation modules within the encoders. It consists of \textbf{a scaling vector $a$}, initialized as an \textbf{all-one} vector, and \textbf{a shifting vector $b$}, initialized as an \textbf{all-zero} vector~\cite{SSF}. The agent layer can be appended after various positions:
\begin{equation}
     {\rm AL} \circ {\rm OP}(x) = {\rm OP}(x) \cdot \Lambda(a) + b,
\end{equation}
where ${\rm OP}$ is a specific computation module, $\Lambda(a)$ represents the diagonal matrix with the vector $a$ as its diagonal elements, $\cdot$ denotes the matrix multiplication. In a transformer block, the agent layer can be positioned after the LayerNorm (LN) layer (\textbf{\textit{Position-1}})  to capture updates related to the parameters of LN. Similarly, it can be placed after the multi-head self-attention (MHSA) layer (\textbf{\textit{Position-2}}) to monitor updates to the output weight matrix $W^O$, or after the MLP layer (\textbf{\textit{Position-3}}) to track changes in the second linear layer, $W_{mlp}^2$. Beyond the confines of transformer blocks, the agent layer can also be appended after the final LN layer (\textbf{\textit{Position-4}}) and the last projection layer $W_{proj}$ (\textbf{\textit{Position-5}}), effectively capturing the relevant updates. 
In summary, the agent layer is specifically designed to track updates for \textbf{five positions} from \textbf{two types} of layers: the LN layer and the linear layer. For illustrative purposes, we will focus on the text encoder (\ie, $a, b \in \mathbb{R}^{d_t}$) to demonstrate how updates occur in the LN and linear layers.

\textbf{LN Layer:} The LN operation is formulated as:
\begin{equation}
    {\rm LN}(x) = \frac{x-\mu}{\sigma} \odot \gamma + \beta,
\end{equation}
where $x \! \in \! \mathbb{R}^{N_t \times d_t}$ denotes the input to the LN layer, $\mu, \sigma \! \in \! \mathbb{R}^{N_t} $, $\gamma, \beta \! \in \! \mathbb{R}^{d_t}$. The agent layer, appended after the LN layer, can be formulated as:
\begin{equation}
 {\rm AL} \circ {\rm LN}(x) = \frac{x-\mu}{\sigma} \odot \underbrace{\gamma \odot \underline{a}} _{\underline{\gamma}}+  \underbrace{\beta \odot \underline{a} + \underline{b}}_{\underline{\beta}}, 
\label{eq: ln_layer_agent}
\end{equation}
where $\underline{\text{underline}}$ indicates the trainable component, and $\odot$ represents the Hadamard product. As shown in \textbf{Eq.~\eqref{eq: ln_layer_agent}}, the update of the agent layer can approximately correspond to the updates of $\gamma$ and $\beta$ in the LN layer.

\textbf{Linear Layer:} The linear layer is formulated as:
\begin{equation}
     {\rm LiL}(x)  = x\cdot W^{\top} + bias,
\end{equation}
where $x \! \in \! \mathbb{R}^{N_t \times d_a}$ represents the input, with $d_a$ can either equal to $d_t$ or the intermediate dimension ( $4\cdot d_t$) of the MLP layer within the transformer block. $W \! \in \! \mathbb{R}^{d_t \times d_a}$ denotes the weight matrix, and $bias \in \mathbb{R}^{d_t}$ represents the bias. The agent layer, appended after the linear layer, is expressed as:
\begin{equation}
      {\rm AL} \circ {\rm LiL}(x)  =  x\cdot {\underbrace{(\Lambda(\underline{a}) \cdot W)}_{\underline{W}}}^{\top}  + \underbrace{bias \odot \underline{a} + \underline{b}}_{\underline{bias}}.
      \label{eq: linear_layer_agent}
\end{equation}
Due to the left multiplication by $\Lambda(a)$ applied to $W$, the updates of $W$ are row-wise.

During inference, based on \textbf{Eq.~\eqref{eq: ln_layer_agent}} and \textbf{Eq.~\eqref{eq: linear_layer_agent}}, the agent layers can be seamlessly integrated into the original foundation model, \textbf{eliminating additional inference latency.}

\vspace{-0.25\baselineskip}
\subsection{Agent Layer Coupling} 
We define the text agent layer as $  {\rm AL_{text}} \! = \! \{a_t, b_t  \!  \in  \! \mathbb{R}^{d_t}\}$, and the image agent layer at the same location in the image encoder as $ {\rm AL_{image}} \! = \! \{a_v, b_v \! \in \! \mathbb{R}^{d_v}\}$ or $ {\rm AL_{image}} \! = \! \{a_v, b_v \! \in \! \mathbb{R}^{d_t}\}$ (when the agent layer is appended after the final projection layer, \ie, position-5). These agent layers can be trained independently, and we term such a design as \textit{independent vision-language updating }(IVLU). To enhance the synergy between the vision and language agent layers, we introduce a multi-modal agent layer coupling approach.
Specifically, the scaling vector $a_v$  in the vision agent layer is integrated with $a_t$ via a bottleneck-based language-to-vision projection, acting as a bridge function that facilitates gradient exchange and promotes aligned updates across the modalities:
\begin{equation}
    \bar{a}_v = a_v + W_{up}\cdot W_{down} \cdot a_t, \label{eq: text2image}
\end{equation}
where $W_{up} \! \in \! \mathbb{R}^{d_v \times r}$ or  $ W_{up} \! \in \! \mathbb{R}^{d_t \times r}$  and $W_{down} \! \in \!\mathbb{R}^{r \times d_t}$, with $r$ representing the rank of the bridge function. Following the initialization method in LoRA~\cite{LoRA}, $W_{down}$ is initialized with random Gaussian values, i.e., $W_{down} \! \sim \! \mathcal{N}(0,\sigma_t^2)$, with $\sigma_t \! = \! \frac{1}{\sqrt{d_t}}$, while $W_{up}$ is initialized to zeros. 
We term this design the \textbf{M}ulti-\textbf{M}od\textbf{a}l \textbf{I}nteractive \textbf{A}gent \textbf{L}ayer (MAIL), where $a_v$ is replaced by  $\bar{a}_v$ in the image agent layer.

\begin{algorithm}[t]
\caption{Pseudocode of MAIL++ in a PyTorch-like style.}
\label{alg: mailpp_code}
\definecolor{codeblue}{rgb}{0.25,0.5,0.5}
\lstset{
  basicstyle=\ttfamily\small,
  backgroundcolor=\color{white},
  columns=fullflexible,
  breaklines=true,
  keepspaces=true,
  captionpos=b,
  commentstyle=\fontsize{9pt}{9pt}\color{codeblue},
  keywordstyle=\fontsize{9pt}{9pt},
  literate={_}{\_}1
}
\vskip -0.1in
\begin{lstlisting}[language=python]
# x: input feature ; y: output feature
# flag: indicates where the input feature x from.
# imageAgent: the agent layer for the image encoder
# textAgent: the agent layer for the text encoder
# W_text, W_visual: text and image bridge functions
# m_a: the meta scaling layer
if flag==''text'':
    a = textAgent.a + m_a @ W_text.down @ W_text.up
    y = x * a + textAgent.b
else:
    a = imageAgent.a + m_a @ W_visual.down @ W_visual.up
    y = x * a + visionAgent.b
\end{lstlisting}
\vskip -0.1in
\end{algorithm}

\vspace{-0.25\baselineskip}
\subsection{Bi-directional Agent Layer Coupling} 
In \textbf{Eq.~\eqref{eq: text2image}}, cross-modal interaction is realized through a single text-to-image bridge, resulting in a one-way dependency between the two encoders. Under this design, gradients originating from the textual branch are propagated to the visual branch, whereas the reverse direction is not explicitly modeled, leading to asymmetric update dynamics during optimization.

To overcome this asymmetry, MAIL++ introduces a bidirectional bridging architecture that enables reciprocal information exchange between the visual and textual streams. As illustrated in \textbf{Fig.~\ref{fig: mailvsmail++}-b}, this is achieved by incorporating a meta-scaling vector alongside modality-specific meta-text and meta-image bridges, which jointly regulate the magnitude and direction of cross-modal interactions:
\begin{subequations}
    \begin{align}
        & \hat{a}_v = a_v + W_{up}^{v}\cdot W_{down}^{v} \cdot a_m, \label{eq: }\\
        & \hat{a}_t = a_t + W_{up}^{t}\cdot W_{down}^{t} \cdot a_m, \label{eq.12B}
    \end{align} 
\end{subequations}
where $a_m \in \mathbb{R}^{d_m}$ denotes the meta-scaling vector with $d_m$ denotes the dimension of the meta scaling layer. $W_{up}^v \! \in \! \mathbb{R}^{d_v \times r}$  or $W_{up}^v \! \in \! \mathbb{R}^{d_t \times r}$, $W_{down}^v \! \in \!\mathbb{R}^{r \times d_m}$, $W_{up}^t \! \in \! \mathbb{R}^{d_t \times r}$, and $W_{down}^t \! \in \!\mathbb{R}^{r \times d_m}$, with $r$ representing the rank of the bridge function.  $\hat{a}_v$ and  $\hat{a}_t$ will replace in $a_v$ in the image agent layer and $a_t$ in the text agent layer, respectively.  
This formulation allows the coupling strength to be adaptively learned from data, facilitating more balanced optimization across modalities. \textbf{Alg.~\ref{alg: mailpp_code}} provides the pseudo-code for MAIL++ in a PyTorch-like style. With just a few lines, MAIL++ can significantly boost performance in a plug-and-play manner.

\subsection{Training Losses}

Given a mini-batch of size $B$, we adopt an image–text matching objective formulated as a cross-entropy loss:
\begin{equation}
      \mathcal{L}_{ce} = \frac{1}{B} \sum_{j=1}^B -y_j \log \frac{ \exp(sim(\mathcal{T}^a(t_{y_j}),  \mathcal{V}^a(I_j)))}{\sum_{i=1}^C \exp(sim(\mathcal{T}^a(t_i), \mathcal{V}^a(I_j))) },
\end{equation}
where $I_j$ denotes the $j$-th image in the batch and $y_j$ is its corresponding class label. Here, $\mathcal{T}^a(\cdot)$ and $\mathcal{V}^a(\cdot)$ denote the text and image encoders augmented with agent layers, respectively, and $sim(\cdot,\cdot)$ represents cosine similarity.

To preserve the generalization capability of pretrained CLIP representations, we further introduce a regularization objective that explicitly constrains the agent-updated features to remain close to the frozen CLIP features:
\begin{subequations}
    \begin{align}
        & \mathcal{L}_{reg}^v = 1 - \frac{1}{B} \sum_{j=1}^B sim(\mathcal{V}^a(I_j) ,  \mathcal{V}(I_j)),\\
        & \mathcal{L}_{reg}^t = 1 - \frac{1}{C} \sum_{i=1}^C sim(\mathcal{T}^a(t_i) ,  \mathcal{T}(t_i)), 
    \end{align} 
\end{subequations}
where $\mathcal{V}(\cdot)$ and $\mathcal{T}(\cdot)$ denote the frozen image and text encoders of CLIP, respectively.
The final loss function is
\begin{equation}
    \mathcal{L} =  \mathcal{L}_{ce} + \lambda \cdot (\mathcal{L}_{reg}^v +  \mathcal{L}_{reg}^t)
\end{equation}
where $\lambda$ is a trade-off hyper-parameter.

\section{Experiments}
In this section, we evaluate the effectiveness of our proposed methods across two categories of few-shot tasks. \textbf{The first category} focuses on classification, including \ding{182}~Base-to-Novel evaluation, \ding{183}~Cross-Dataset evaluation, and \ding{184}~Domain Generalization evaluation. \textbf{The second category} pertains to few-shot retrieval, encompassing \ding{182}~FS-UCDR, \ding{183}~FS-U$^C$CDR, and \ding{184}~FS-U$^D$CDR, which assess the model’s cross-domain retrieval ability across unssen domains and classes.


\begin{table*}[t]
\centering
\caption{Base-to-Novel (16-shot) evaluation results (\%) across 11 datasets. For a fair comparison, methods that utilize LLMs or are distilled from larger CLIP backbones are excluded. The best performance under FS-UCDR is marked as \textbf{bold} and the second best performance is marked as \underline{underline}, while scores from our methods are highlighted with a \color{Purple}light purple background.}
\vspace{-0.3\baselineskip}
\label{tab: base_to_novel}
\renewcommand\arraystretch{0.92} 
\setlength{\tabcolsep}{7.0pt}
\resizebox{0.9\linewidth}{!}{
    \begin{tabular}{l|ccc|ccc|ccc|ccc}
    \toprule[1.1pt]
    \textbf{\multirow{2}{*}[-0.5ex]{Methods}} & \multicolumn{3}{c|}{\textit{\textbf{Average}}} & \multicolumn{3}{c|}{ImageNet} & \multicolumn{3}{c|}{Caltech101} & \multicolumn{3}{c}{OxfordPets} \\
    & Base & Novel & HM & Base & Novel & HM & Base & Novel & HM & Base & Novel & HM \\
    \midrule
    {CLIP}\textcolor{gray}{\scriptsize{ [ICML'21]}} &
    $69.34$ & $74.22$ & $71.70$ & $72.43$ & $68.14$ & $70.22$ & $96.84$ & $94.00$ & $95.40$ & $91.17$ & $97.26$ & $94.12$ \\
    CoOp\textcolor{gray}{\scriptsize{ [IJCV'22]}}&
    $82.69$ & $63.22$ & $71.66$ & $76.47$ & $67.88$ & $71.92$ & $98.00$ & $89.81$ & $93.73$ & $93.67$ & $95.29$ & $94.47$ \\
    CoOpOp\textcolor{gray}{\scriptsize{ [CVPR'22]}}&
    $80.47$ & $71.69$ & $75.83$ & $75.98$ & $70.43$ & $73.10$ & $97.96$ & $93.81$ & $95.84$ & $95.20$ & $97.69$ & $96.43$ \\
    KgCoOp\textcolor{gray}{\scriptsize{ [CVPR'23]}}&
    $80.73$ & $73.60$ & $77.00$ & $75.83$ & $69.96$ & $72.78$ & $97.72$ & $94.39$ & $96.03$ & $94.65$ & $97.76$ & $96.18$ \\
    MaPLe\textcolor{gray}{\scriptsize{ [CVPR'23]}} &
    $82.28$ & $75.14$ & $78.55$ & $76.66$ & $70.54$ & $73.47$ & $97.74$ & $94.36$ & $96.02$ & $95.43$ & $97.76$ & $96.58$ \\
    PromptSRC\textcolor{gray}{\scriptsize{ [ICCV'23]}} & 
    $84.26$ & $76.10$ & $79.97$ & $77.60$ & $70.73$ & $74.01$ & $98.10$ & $94.03$ & $96.02$ & $95.33$ & $97.30$ & $96.30$ \\
    TCP\textcolor{gray}{\scriptsize{ [CVPR'24]}} &
    $84.13$ & $75.36$ & $79.51$ & $77.27$ & $69.87$ & $73.38$ & $98.23$ & $94.67$ & $96.42$ & $94.67$ & $97.20$ & $95.92$ \\ 
    MMA\textcolor{gray}{\scriptsize{ [CVPR'24]}} &
    $83.20$ & $76.80$ & $79.87$ & $77.31$ & $71.00$ & $74.02$ & $98.40$ & $94.00$ & $96.15$ & $95.40$ & $98.07$ & $96.72$ \\ 
    DeKg\textcolor{gray}{\scriptsize{ [ICLR'25]}} & 
    $84.96$ & $76.38$ & $80.44$ & $77.40$ & $69.20$ & $73.07$ & $98.64$ & $95.20$ & \underline{$96.89$} & $94.47$ & $97.76$ & $96.09$ \\ 
    MMRL\textcolor{gray}{\scriptsize{ [CVPR'25]}} &
    $85.68$ & $77.16$ & $81.20$ & $77.90$ & \underline{$71.30$} & \underline{$74.45$} & \textbf{98.97} & $94.50$ & $96.68$ & \underline{$95.90$} & $97.60$ & $96.74$ \\
    BIP-D\textcolor{gray}{\scriptsize{ [TPAMI'25]}} &
    \textbf{85.95} & $75.79$ & $80.55$ & \textbf{78.27} & $69.90$ & $73.85$ & $98.77$ & $94.47$ & $96.57$ & $95.30$ & $97.37$ & $96.32$ \\
    FCPrompt-G\textcolor{gray}{\scriptsize{ [TPAMI'25]}} &
    $85.28$ & $75.72$ & $80.22$ & $77.33$ & $71.11$ & $74.09$ & $98.41$ & $94.07$ & $96.19$ & $96.40$ & $97.89$ & \textbf{97.14}\\  
    \midrule
    \cellcolor{lightPurple}\textbf{MAIL}\textcolor{gray}{\scriptsize{ [Ours, NeurIPS'25]}} &
    \cellcolor{lightPurple}$85.19$ &
    \cellcolor{lightPurple}$77.39$ &
    \cellcolor{lightPurple}$81.10$ &
    \cellcolor{lightPurple}$77.92$ &
    \cellcolor{lightPurple}$71.22$ &
    \cellcolor{lightPurple}$74.42$ &
    \cellcolor{lightPurple}$98.34$ &
    \cellcolor{lightPurple}\underline{$95.36$} &
    \cellcolor{lightPurple}$96.83$ &
    \cellcolor{lightPurple}$95.50$ &
    \cellcolor{lightPurple}$97.97$ &
    \cellcolor{lightPurple}$96.72$ \\
    \cellcolor{lightPurple}\textbf{MAIL++}\textcolor{gray}{\scriptsize{ [Ours]}} &
    \cellcolor{lightPurple}\underline{$85.69$}&
    \cellcolor{lightPurple}\textbf{78.73}&
    \cellcolor{lightPurple}\textbf{82.06}&
    \cellcolor{lightPurple}\underline{$78.10$} & 
    \cellcolor{lightPurple}\textbf{71.87} &
    \cellcolor{lightPurple}\textbf{74.86} &
    \cellcolor{lightPurple}\underline{$98.80$} &
    \cellcolor{lightPurple}\textbf{95.53} &
    \cellcolor{lightPurple}\textbf{97.14} &
    \cellcolor{lightPurple}\textbf{96.10} &
    \cellcolor{lightPurple}\textbf{98.20} &
    \cellcolor{lightPurple}\textbf{97.14} \\
    \midrule 
    \multirow{2}{*}{\textbf{Methods}} & \multicolumn{3}{c|}{StanfordCars} & \multicolumn{3}{c|}{Flowers102} &
    \multicolumn{3}{c|}{Food101}      &  \multicolumn{3}{c}{FGVCAircraft} \\ &
    Base & Novel & HM & Base & Novel & HM & Base & Novel & HM & Base & Novel & HM \\
    \midrule
    CLIP\textcolor{gray}{\scriptsize{ [ICML'21]}} &
    $63.37$ & $74.89$ & $68.65$ & $72.08$ & \underline{$77.80$} & $74.83$ & $90.10$ & $91.22$ & $90.66$ & $27.19$ & $36.29$ & $31.09$ \\
    CoOp\textcolor{gray}{\scriptsize{ [IJCV'22]}} & 
    $78.12$ & $60.40$ & $68.13$ & $97.60$ & $59.67$ & $74.06$ & $88.33$ & $82.26$ & $85.19$ & $40.44$ & $22.30$ & $28.75$ \\
    CoOpOp\textcolor{gray}{\scriptsize{ [CVPR'22]}} & 
    $70.49$ & $73.59$ & $72.01$ & $94.87$ & $71.75$ & $81.71$ & $90.70$ & $91.29$ & $90.99$ & $33.41$ & $23.71$ & $27.74$ \\
    KgCoOp\textcolor{gray}{\scriptsize{ [CVPR'22]}} &
    $71.76$ & \underline{$75.04$} & $73.36$ & $95.00$ & $74.73$ & $83.65$ & $90.50$ & $91.70$ & $91.09$ & $36.21$ & $33.55$ & $34.83$ \\
    MaPLe\textcolor{gray}{\scriptsize{ [CVPR'23]}} & 
    $72.94$ & $74.00$ & $73.47$ & $95.92$ & $72.46$ & $82.56$ & $90.71$ & $92.05$ & $91.38$ & $37.44$ & $35.61$ & $36.50$ \\
    PromptSRC\textcolor{gray}{\scriptsize{ [ICCV'23]}} &
    $78.27$ & $74.97$ & $76.58$ & $98.07$ & $76.50$ & $85.95$ & $90.67$ & $91.53$ & $91.10$ & $42.73$ & \textbf{37.87} & $40.15$ \\
    MMA\textcolor{gray}{\scriptsize{ [CVPR'24]}} &
    $78.50$ & $73.10$ & $75.70$ & $97.77$ & $75.93$ & $85.48$ & $90.13$ & $91.30$ & $90.71$ & $40.57$ & $36.33$ & $38.33$ \\
    TCP\textcolor{gray}{\scriptsize{ [CVPR'24]}} &
    $80.80$ & $74.13$ & $77.32$ & $97.73$ & $75.57$ & $85.23$ & $90.57$ & $91.37$ & $90.97$ & $41.97$ & $34.43$ & $37.83$ \\
    DeKg\textcolor{gray}{\scriptsize{ [ICLR'25]}} &
    $81.18$ & $74.75$ & $77.83$ & $98.58$ & $75.18$ & $85.30$ & $90.73$ & $91.55$ & $91.14$ & $45.20$ & $35.09$ & $39.51$ \\ 
    MMRL\textcolor{gray}{\scriptsize{ [CVPR'25]}} &
    $81.30$ & \textbf{75.07} & \underline{$78.06$} & \textbf{98.97} & $77.27$ & \underline{$86.78$} & $90.57$ & $91.50$ & $91.03$ & $46.30$ & $37.03$ & $41.15$ \\
    BIP-D\textcolor{gray}{\scriptsize{ [TPAMI'25]}} &
    \textbf{83.27} & $73.00$ & $77.80$ & $98.30$ & $75.90$ & $85.66$ & $90.73$ & $91.23$ & $90.98$ & \underline{$48.83$} & $35.90$ & $41.38$ \\
    FCPrompt-G\textcolor{gray}{\scriptsize{ [TPAMI'25]}} &
    $80.87$ & $73.03$ & $76.75$ & \underline{$98.32$} & $74.82$ & $84.98$ & \underline{$90.76$} & \textbf{92.19} & \underline{$91.47$} & $43.68$ & $36.97$ & $40.05$\\  
    \midrule
    \cellcolor{lightPurple}\textbf{MAIL}\textcolor{gray}{\scriptsize{ [Ours, NeurIPS'25]}} &
    \cellcolor{lightPurple}$82.27$ &
    \cellcolor{lightPurple}$72.03$ &
    \cellcolor{lightPurple}$76.81$ &
    \cellcolor{lightPurple}$98.20$ &
    \cellcolor{lightPurple}$75.27$ &
    \cellcolor{lightPurple}$85.22$ &
    \cellcolor{lightPurple}$90.54$ &
    \cellcolor{lightPurple}$91.77$ &
    \cellcolor{lightPurple}$91.15$ &
    \cellcolor{lightPurple}$47.80$ &
    \cellcolor{lightPurple}$36.27$ &
    \cellcolor{lightPurple}$41.24$ \\ 
    \cellcolor{lightPurple}\textbf{MAIL++}\textcolor{gray}{\scriptsize{ [Ours]}} &
    \cellcolor{lightPurple}\underline{$82.50$}&
    \cellcolor{lightPurple}$74.47$&
    \cellcolor{lightPurple}\textbf{78.28}&
    \cellcolor{lightPurple}$98.23$&
    \cellcolor{lightPurple}\textbf{77.97}&
    \cellcolor{lightPurple}\textbf{86.94}&
    \cellcolor{lightPurple}\textbf{91.10}&
    \cellcolor{lightPurple}\underline{$91.97$}&
    \cellcolor{lightPurple}\textbf{91.53}&
    \cellcolor{lightPurple}\textbf{48.87}&
    \cellcolor{lightPurple}\underline{$37.40$}&
    \cellcolor{lightPurple}\textbf{42.37}\\
    \midrule
    \multirow{2}{*}{\textbf{Methods}} & \multicolumn{3}{c|}{SUN397} & \multicolumn{3}{c|}{DTD} & \multicolumn{3}{c|}{EuroSAT} & \multicolumn{3}{c}{UCF101} \\
    & Base & Novel & HM & Base & Novel & HM & Base & Novel & HM & Base & Novel & HM \\ 
    \midrule
    CLIP\textcolor{gray}{\scriptsize{ [ICML'21]}} &
    $69.36$ & $75.35$ & $72.23$ & $53.24$ & $59.90$ & $56.37$ & $56.48$ & $64.05$ & $60.03$ & $70.53$ & $77.50$ & $73.85$ \\
    CoOp\textcolor{gray}{\scriptsize{ [IJCV'22]}} & 
    $80.60$ & $65.89$ & $72.51$ & $79.44$ & $41.18$ & $54.24$ & $92.19$ & $54.74$ & $68.69$ & $84.69$ & $56.05$ & $67.46$ \\
    CoOpOp\textcolor{gray}{\scriptsize{ [CVPR'22]}} & 
    $79.74$ & $76.86$ & $78.27$ & $77.01$ & $56.00$ & $64.85$ & $87.49$ & $60.04$ & $71.21$ & $82.33$ & $73.45$ & $77.64$ \\
    KgCoOp\textcolor{gray}{\scriptsize{ [CVPR'22]}} &
    $80.29$ & $76.53$ & $78.36$ & $77.55$ & $54.99$ & $64.35$ & $85.64$ & $64.34$ & $73.48$ & $82.89$ & $76.67$ & $79.65$ \\
    MaPLe\textcolor{gray}{\scriptsize{ [CVPR'23]}} &
    $80.82$ & $78.70$ & $79.75$ & $80.36$ & $59.18$ & $68.16$ & $94.07$ & $73.23$ & $82.35$ & $83.00$ & $78.66$ & $80.77$ \\
    PromptSRC\textcolor{gray}{\scriptsize{ [ICCV'23]}} &
    $82.67$ & $78.47$ & $80.52$ & $83.37$ & $62.97$ & $71.75$ & $92.90$ & $73.90$ & $82.32$ & $87.10$ & $78.80$ & $82.74$ \\
    MMA\textcolor{gray}{\scriptsize{ [CVPR'24]}} &
    $82.27$ & $78.57$ & $80.38$ & $83.20$ & $65.63$ & $73.38$ & $85.46$ & $82.34$ & $83.87$ & $86.23$ & $80.03$ & $82.20$ \\ 
    TCP\textcolor{gray}{\scriptsize{ [CVPR'24]}} &
    $82.63$ & $78.20$ & $80.35$ & $82.77$ & $58.07$ & $68.25$ & $91.63$ & $74.73$ & $82.32$ & $87.13$ & $80.77$ & $83.83$ \\
    DeKg\textcolor{gray}{\scriptsize{ [ICLR'25]}} &
    $82.52$ & $78.30$ & $80.35$ & $83.80$ & $59.66$ & $69.70$ & $94.02$ & $81.69$ & $87.42$ & \underline{$88.06$} & \underline{$81.77$} & \textbf{84.80} \\ 
    MMRL\textcolor{gray}{\scriptsize{ [CVPR'25]}} &
    \textbf{83.20} & \underline{$79.30$} & \underline{$81.20$} & \textbf{85.67} & $65.00$ & $73.82$ & \underline{$95.60$} & $80.17$ & $87.21$ & \textbf{88.10} & $80.07$ & $83.89$ \\
    BIP-D\textcolor{gray}{\scriptsize{ [TPAMI'25]}} &
    \underline{$83.07$} & $79.13$ & $81.05$ & \underline{$85.43$} & $58.77$ & $69.63$ & \textbf{95.70} & $78.00$ & $85.95$ & $87.77$ & $80.00$ & $83.70$ \\
    FCPrompt-G\textcolor{gray}{\scriptsize{ [TPAMI'25]}} &
    $82.93$ & $79.06$ & $80.95$ & $84.61$ & $60.31$ & $70.42$ & $96.80$ & $75.21$ & $84.65$ & $87.99$ & $78.28$ & $82.85$\\  
    \midrule
    \cellcolor{lightPurple}\textbf{MAIL}\textcolor{gray}{\scriptsize{ [Ours, NeurIPS'25]}} &
    \cellcolor{lightPurple}$82.50$&
    \cellcolor{lightPurple}$78.70$&
    \cellcolor{lightPurple}$80.56$&
    \cellcolor{lightPurple}$83.15$&
    \cellcolor{lightPurple}\underline{$67.39$}&
    \cellcolor{lightPurple}\underline{$74.45$}&
    \cellcolor{lightPurple}$93.50$&
    \cellcolor{lightPurple}\underline{$85.11$}&
    \cellcolor{lightPurple}\underline{$89.11$}&
    \cellcolor{lightPurple}$87.34$&
    \cellcolor{lightPurple}$80.22$&
    \cellcolor{lightPurple}$83.63$\\ 
    \cellcolor{lightPurple}\textbf{MAIL++}\textcolor{gray}{\scriptsize{ [Ours]}} &
    \cellcolor{lightPurple}$82.73$&
    \cellcolor{lightPurple}\textbf{79.83}&
    \cellcolor{lightPurple}\textbf{81.25}&
    \cellcolor{lightPurple}$83.13$&
    \cellcolor{lightPurple}\textbf{67.93}&
    \cellcolor{lightPurple}\textbf{74.77}&
    \cellcolor{lightPurple}$95.17$&
    \cellcolor{lightPurple}\textbf{89.00}&
    \cellcolor{lightPurple}\textbf{91.98}&
    \cellcolor{lightPurple}$87.90$&
    \cellcolor{lightPurple}\textbf{81.87}&
    \cellcolor{lightPurple}\underline{$84.78$}\\
    
    \bottomrule
    \end{tabular}
}
\vspace{-0.6\baselineskip}
\end{table*}

\begin{table*}[t]
\centering
\caption{Comparison of MAIL with previous state-of-the-art methods on cross-dataset evaluation.}
\vspace{-0.3\baselineskip}
\label{tab: cross_dataset}
\renewcommand\arraystretch{0.92} 
\setlength{\tabcolsep}{5.5pt}
\resizebox{0.9\textwidth}{!}{
    \begin{tabular}{lc|ccccccccccc}
      \toprule
      & Source & \multicolumn{11}{c}{Target} \\ 
      \cmidrule(l){2-13} 
      &
      \rotatebox{60}{ImageNet} &
      \rotatebox{60}{Average} &
      \rotatebox{60}{Caltech101} &
      \rotatebox{60}{OxfordPets} &
      \rotatebox{60}{StanfordCars} &
      \rotatebox{60}{Flowers101} &
      \rotatebox{60}{Food101} &
      \rotatebox{60}{FGVCAircraft} &
      \rotatebox{60}{SUN397} &
      \rotatebox{60}{DTD} &
      \rotatebox{60}{EuroSAT} &
      \rotatebox{60}{UCF101} \\ \cmidrule(l){2-13} 
      CoOp\textcolor{gray}{\scriptsize{ [IJCV'22]}}&
      $71.51$ & $63.88$ & $93.70$ & $89.14$ & $64.51$ & $68.71$ & $85.30$ & $18.47$ & $64.15$ & $41.92$ & $46.39$ & $66.55$ \\
      CoOpOp\textcolor{gray}{\scriptsize{ [CVPR'22]}} &
      $71.02$ & $65.74$ & $94.43$ & $90.14$ & $65.32$ & $71.88$ & $86.06$ & $22.94$ & $67.36$ & $45.73$ & $45.37$ & $68.21$ \\
      MaPLe\textcolor{gray}{\scriptsize{ [CVPR'23]}} &
      $70.72$ & $66.30$ & $93.53$ & $90.49$ & $65.57$ & $72.23$ & $86.20$ & $24.74$ & $67.01$ & $46.49$ & $48.06$ & $68.69$ \\
      PromptSRC\textcolor{gray}{\scriptsize{ [ICCV'23]}} &
      $71.27$ & $65.81$ & $93.60$ & $90.25$ & $65.70$ & $70.25$ & $86.15$ & $23.90$ & $67.10$ & \textbf{46.87} & $45.50$ & $68.75$ \\
      MMA\textcolor{gray}{\scriptsize{ [CVPR'24]}} &
      $71.00$ & $66.61$ & $93.80$ & $90.30$ & $66.13$ & $72.07$ & $86.12$ & $25.33$ & \textbf{68.17} & \underline{$46.57$} & $49.24$ & $68.32$ \\ 
      TCP\textcolor{gray}{\scriptsize{ [CVPR'24]}} &
      $71.40$ & $66.29$ & $93.97$ & $91.25$ & $64.69$ & $71.21$ & $86.69$ & $23.45$ & $67.15$ & $44.35$ & $51.45$ & $68.73$ \\ 
      DeKg\textcolor{gray}{\scriptsize{ [ICLR'25]}} &
      \textbf{72.33} & $66.64$ &  \textbf{94.73} & $90.02$ & $65.49$ & $72.39$ & \underline{$86.59$} & $25.05$ & $67.19$ & $44.47$ & $51.37$ & $68.78$ \\
      MMRL\textcolor{gray}{\scriptsize{ [CVPR'25]}} 
      & $72.03$ & \underline{$67.25$} & \underline{$94.67$} & \textbf{91.43} & $66.10$ & \underline{$72.77$} & $86.40$ & $26.30$
      & $67.57$ & $45.90$ & \underline{$53.10$} & $68.27$ \\
      BIP-D\textcolor{gray}{\scriptsize{ [TPAMI'25]}}
      & $70.83$ & $66.57$ & $93.93$ & $90.13$ & $65.53$ & $71.43$ & $86.27$ & \underline{$26.47$} & $67.23$ & $46.00$ & $48.90$ & \textbf{69.83} \\
      \midrule
      \cellcolor{lightPurple}\textbf{MAIL}\textcolor{gray}{\scriptsize{ [Ours, NeurIPS'25]}} &
      \cellcolor{lightPurple}\underline{$72.10$} &
      \cellcolor{lightPurple}$67.02$ &
      \cellcolor{lightPurple}\textbf{94.73} &
      \cellcolor{lightPurple}\underline{$91.37$} &
      \cellcolor{lightPurple}\underline{$66.63$} &
      \cellcolor{lightPurple}$71.47$ &
      \cellcolor{lightPurple}$86.33$ &
      \cellcolor{lightPurple}$25.27$ &
      \cellcolor{lightPurple}$67.30$ &
      \cellcolor{lightPurple}$45.47$ &
      \cellcolor{lightPurple}$52.80$ &
      \cellcolor{lightPurple}$68.87$ \\ 
      \cellcolor{lightPurple}\textbf{MAIL++}\textcolor{gray}{\scriptsize{ [Ours]}} &
      \cellcolor{lightPurple}$72.03$&
      \cellcolor{lightPurple}\textbf{67.98}&
      \cellcolor{lightPurple}$94.60$&
      \cellcolor{lightPurple}$91.20$&
      \cellcolor{lightPurple}\textbf{66.70}&
      \cellcolor{lightPurple}\textbf{72.87}&
      \cellcolor{lightPurple}\textbf{86.63}&
      \cellcolor{lightPurple}\textbf{26.87}&
      \cellcolor{lightPurple}\underline{$67.87$}&
      \cellcolor{lightPurple}$46.37$&
      \cellcolor{lightPurple}\textbf{57.53}&
      \cellcolor{lightPurple}\underline{$69.17$}\\ 
      \bottomrule
    \end{tabular}
}
\vspace{-0.6\baselineskip}
\end{table*}

\begin{table}[t]
\centering
\caption{Comparison of MAIL++ with previous state-of-the-art methods on Domain Generalization. Results of DeKg are derived from the OpenReview comment section.}
\vspace{-0.3\baselineskip}
\renewcommand\arraystretch{0.92} 
\setlength{\tabcolsep}{5.0pt}
\label{tab: domain_generalization}
\resizebox{0.9\columnwidth}{!}{
    \begin{tabular}{lc|ccccc}
 
    \toprule
    & Source   & \multicolumn{5}{c}{Target}    \\ 
    \cmidrule(l){2-7} 
    & ImageNet & Avg. &-V2   & -S    & -A    & -R \\ \cmidrule(l){2-7} 
    CLIP   & $66.73$ & $57.17$   & $60.83$ & $46.15$ & $47.77$ & $73.96$ \\
    CoOp  & $71.51$ & $59.27$  & $64.20$ & $47.99$ & $49.71$ & $75.21$ \\
    CoOpOp & $71.02$ & $59.91$   & $64.07$ & $48.75$ & $50.63$ & $76.18$ \\
    MaPLe  & $70.72$ &  $60.28$ & $64.07$ & $49.15$ & $50.90$ & $76.98$ \\
    PromptSRC & $71.27$   &$60.65$ & $64.35$ & $49.55$ & $50.90$  & \underline{$77.80$} \\
    MMA     & $71.00$ &$60.47$  & $64.33$ & $49.13$ & \textbf{51.12} & $77.32$ \\
    TCP     & $71.20$ & $60.51$ & $64.60$ & $49.50$ & $51.20$ & $76.73$ \\ 
    DeKg  & \textbf{72.33}	& $59.89$& $64.31$ & $48.38$ & $50.51$ & $76.37$ \\
    MMRL    & $72.03$ & $60.66$ & $64.47$ & $49.17$ & $51.20$ & $77.80$ \\ 
    BIP-D  & $70.80$ & $60.50$ & $64.40$ & $50.00$ & $50.30$ & $77.30$ \\ 
    \midrule
    \cellcolor{lightPurple}\textbf{MAIL}       
     & \cellcolor{lightPurple}\underline{$72.10$} 
     & \cellcolor{lightPurple}\underline{$60.68$}
     & \cellcolor{lightPurple}\underline{$64.50$} 
     & \cellcolor{lightPurple}\underline{$49.67$}  
     & \cellcolor{lightPurple}$50.70$ 
     & \cellcolor{lightPurple}\underline{$77.80$} \\
    \cellcolor{lightPurple}\textbf{MAIL++}      
     & \cellcolor{lightPurple}$72.03$
     & \cellcolor{lightPurple}\textbf{61.26} 
     & \cellcolor{lightPurple}\textbf{65.03}
     & \cellcolor{lightPurple}\textbf{50.40}
     & \cellcolor{lightPurple}\underline{$51.10$}
     & \cellcolor{lightPurple}\textbf{78.50} \\
    \bottomrule
    \end{tabular}
}
\vspace{-1.0\baselineskip}
\end{table}

\begin{table*}[t]
\centering
\caption{FS-UCDR (2-shot) evaluation results (\%) on DomainNet. * denotes the results are obtained using the full training data, i.e., the UCDR results.}
\vspace{-0.3\baselineskip}
\renewcommand\arraystretch{0.92} 
\setlength{\tabcolsep}{4.3pt}
\resizebox{0.9\linewidth}{!}{
    \begin{tabular}{l|cccc|cccc|cccc}
\toprule[1.1pt]
\textbf{ \multirow{3}{*}[-0.5ex]{\centering Methods}}  & \multicolumn{4}{c|}{\textbf{\textit{Sketch}}} & \multicolumn{4}{c|}{\textbf{\textit{Quickdraw}}} & \multicolumn{4}{c}{\textbf{\textit{Painting}}} \\ 
\cmidrule(lr){2-5} \cmidrule(lr){6-9} \cmidrule(lr){10-13} 
& \multicolumn{2}{c}{\textbf{UnseenGallery}} & \multicolumn{2}{c|}{\textbf{MixedGallery}} & \multicolumn{2}{c}{\textbf{UnseenGallery}} & \multicolumn{2}{c|}{\textbf{MixedGallery}} &\multicolumn{2}{c}{\textbf{UnseenGallery}} & \multicolumn{2}{c}{\textbf{MixedGallery}}  \\
& mAP$_{200}$ & Prec$_{200}$ & mAP$_{200}$ & Prec$_{200}$ & mAP$_{200}$ & Prec$_{200}$ & mAP$_{200}$ & Prec$_{200}$ & mAP$_{200}$ & Prec$_{200}$ & mAP$_{200}$ & Prec$_{200}$ \\
\cmidrule(lr){1-1} \cmidrule(lr){2-3} \cmidrule(lr){4-5} \cmidrule(lr){6-7} \cmidrule(lr){8-9} \cmidrule(lr){10-11} \cmidrule(lr){12-13} 

ProS*\textcolor{gray}{\scriptsize{ [CVPR'24]}}
& $64.67$ & $60.01 $ &$58.43$ & $54.63$ & $28.42$ & $25.44$ & $23.18$ & $21.27$ & $75.16$ & $69.55$ & $71.20$ & $66.12$\\
\cmidrule(lr){1-1} \cmidrule(lr){2-3} \cmidrule(lr){4-5} \cmidrule(lr){6-7} \cmidrule(lr){8-9} \cmidrule(lr){10-11} \cmidrule(lr){12-13} 
CLIP\textcolor{gray}{\scriptsize{ [ICML'21]}} 
& $42.20$ & $35.28$ & $36.62$ & $29.79$ & $7.44$  & $5.61$  & $6.00$  & $3.17$  & $61.68$ & $55.07$ & $56.53$ & $50.14$ \\
BitFit\textcolor{gray}{\scriptsize{ [ACL'22]}} 
& $62.68$ & $57.71$ & $55.84$ & $51.52$ & $24.88$ & $21.99$ & $19.22$ & $16.93$ & $73.40$ & $67.39$ & $68.46$ & $62.93$ \\
VPT-D\textcolor{gray}{\scriptsize{ [ECCV'22]}}   
& $57.47$ & $52.58$ & $50.34$ & $45.92$ & $24.26$ & $21.68$ & $18.45$ & $16.77$ & $72.05$ & $65.88$ & $67.25$ & $61.42$ \\ 
AFormer\textcolor{gray}{\scriptsize{ [NeurIPS'22]}}  
& $59.64$ & $53.73$ & $52.91$ & $47.77$ & $23.11$ & $19.86$ & $18.11$ & $15.71$ & $71.08$ & $64.59$ & $66.14$ & $60.17$ \\ 
IVLP\textcolor{gray}{\scriptsize{ [CVPR'23]}}   
& $56.25$ & $51.48$ & $49.37$ & $45.06$ & $21.07$ & $18.97$ & $15.67$ & $14.13$ & $71.47$ & $65.51$ & $66.40$ & $60.86$ \\
IVLA\textcolor{gray}{\scriptsize{ [CVPR'24]}}    
& $60.01$ & $54.16$ & $53.27$ & $48.19$ & $23.96$ & $20.72$ & $18.74$ & $16.38$ & $71.02$ & $64.60$ & $66.10$ & $60.18$ \\
MaPLe\textcolor{gray}{\scriptsize{ [CVPR'23]}}   
& $62.50$ & $57.49$ & $55.69$ & $51.67$ & $27.57$ & $24.93$ & $21.58$ & $19.75$ & $75.02$ & $69.41$ & $70.61$ & $65.38$ \\
MMA\textcolor{gray}{\scriptsize{ [CVPR'24]}}   
& $63.86$ & $58.84$ & $56.74$ & $52.33$ & $27.37$ & $24.28$ & $21.84$ & $19.56$ & $74.08$ & $68.36$ & $69.24$ & $63.92$ \\
\cellcolor{lightPurple}\textbf{\ourmethodCC}\textcolor{gray}{\scriptsize{ [Ours, NeurIPS'25]}}  & 
\cellcolor{lightPurple}$\underline{65.76}$ & \cellcolor{lightPurple}$\underline{61.57}$ & \cellcolor{lightPurple}$\underline{59.05}$ & \cellcolor{lightPurple}$\underline{55.25}$ & \cellcolor{lightPurple}$\underline{29.41}$ & \cellcolor{lightPurple}$\underline{26.95}$ & \cellcolor{lightPurple}$\underline{22.83}$ & \cellcolor{lightPurple}$\underline{21.26}$ & \cellcolor{lightPurple}$\underline{76.05}$ & \cellcolor{lightPurple}$\underline{70.85}$ & \cellcolor{lightPurple}$\underline{71.12}$ & \cellcolor{lightPurple}$\underline{66.44}$ \\
\cellcolor{lightPurple}\textbf{MAIL++}\textcolor{gray}{\scriptsize{ [Ours]}} & 
\cellcolor{lightPurple}$\textbf{68.04}$ & \cellcolor{lightPurple}$\textbf{63.87}$ & \cellcolor{lightPurple}$\textbf{60.90}$ & \cellcolor{lightPurple}$\textbf{57.34}$ & \cellcolor{lightPurple}$\textbf{30.70}$ & \cellcolor{lightPurple}$\textbf{28.50}$ & \cellcolor{lightPurple}$\textbf{24.03}$ & \cellcolor{lightPurple}$\textbf{22.74}$ & \cellcolor{lightPurple}$\textbf{76.55}$ & \cellcolor{lightPurple}$\textbf{71.53}$ & \cellcolor{lightPurple}$\textbf{71.71}$ & \cellcolor{lightPurple}$\textbf{67.24}$ \\
\midrule\midrule
 \textbf{ \multirow{3}{*}[-0.5ex]{Methods}}   & \multicolumn{4}{c|}{\textit{\textbf{Infograph}}} & \multicolumn{4}{c|}{\textit{\textbf{Clipart}}} & \multicolumn{4}{c}{\textit{\textbf{Average}}} \\ 
\cmidrule(lr){2-5} \cmidrule(lr){6-9} \cmidrule(lr){10-13} 
& \multicolumn{2}{c}{\textbf{UnseenGallery}} & \multicolumn{2}{c|}{\textbf{MixedGallery}} & \multicolumn{2}{c}{\textbf{UnseenGallery}} & \multicolumn{2}{c|}{\textbf{MixedGallery}} &\multicolumn{2}{c}{\textbf{UnseenGallery}} & \multicolumn{2}{c}{\textbf{MixedGallery}}  \\
& mAP$_{200}$ & Prec$_{200}$ & mAP$_{200}$ & Prec$_{200}$ & mAP$_{200}$ & Prec$_{200}$ & mAP$_{200}$ & Prec$_{200}$ & mAP$_{200}$ & Prec$_{200}$ & mAP$_{200}$ & Prec$_{200}$ \\
\cmidrule(lr){1-1} \cmidrule(lr){2-3} \cmidrule(lr){4-5} \cmidrule(lr){6-7} \cmidrule(lr){8-9} \cmidrule(lr){10-11} \cmidrule(lr){12-13} 
ProS*\textcolor{gray}{\scriptsize{ [CVPR'24]}}\
& $57.98$ & $54.42$ & $52.19$ & $49.56$ & $76.48$ & $71.86$ & $72.28$ & $68.15$ & $60.52$ & $56.26$ & $55.46$ & $51.95$\\
\cmidrule(lr){1-1} \cmidrule(lr){2-3} \cmidrule(lr){4-5} \cmidrule(lr){6-7} \cmidrule(lr){8-9} \cmidrule(lr){10-11} \cmidrule(lr){12-13} 
CLIP\textcolor{gray}{\scriptsize{ [ICML'21]}} 
& $50.08$ & $44.74$ & $43.75$ & $38.91$ & $60.37$ & $51.30$ & $56.08$ & $46.91$ & $44.35$ & $38.40$ & $39.80$ & $33.78$ \\
BitFit\textcolor{gray}{\scriptsize{ [ACL'22]}} 
& $58.84$ & $55.14$ & $52.32$ & $48.86$ & $75.81$ & $70.27$ & $71.04$ & $65.85$ & $59.12$ & $54.50$ & $53.38$ & $49.22$ \\
VPT-D\textcolor{gray}{\scriptsize{ [ECCV'22]}}
& $56.28$ & $52.18$ & $49.61$ & $45.92$ & $72.96$ & $67.82$ & $68.02$ & $63.04$ & $56.60$ & $52.20$ & $50.73$ & $46.61$ \\
AFormer\textcolor{gray}{\scriptsize{ [NeurIPS'22]}} 
& $59.55$ & $55.00$ & $53.17$ & $49.27$ & $73.19$ & $66.28$ & $68.66$ & $62.10$ & $57.27$ & $51.89$ & $51.79$ & $47.00$ \\ 
IVLP\textcolor{gray}{\scriptsize{ [CVPR'23]}}  
& $58.20$ & $54.07$ & $51.60$ & $47.89$ & $72.28$ & $66.98$ & $67.64$ & $62.47$ & $55.85$ & $51.42$ & $50.13$ & $46.08$ \\
IVLA\textcolor{gray}{\scriptsize{ [CVPR'24]}}  
& $57.39$ & $53.36$ & $50.86$ & $47.22$ & $73.31$ & $66.40$ & $68.75$ & $62.21$ & $57.13$ & $51.84$ & $51.54$ & $46.68$ \\ 
MaPLe\textcolor{gray}{\scriptsize{ [CVPR'23]}}
& $59.45$ & $56.14$ & $53.05$ & $49.95$ & $76.92$ & $72.28$ & $71.98$ & $67.62$ & $60.29$ & $56.05$ & $54.58$ & $50.87$ \\
MMA\textcolor{gray}{\scriptsize{ [CVPR'24]}}
& $59.33$ & $55.01$ & $53.11$ & $49.23$ & $76.95$ & $71.62$ & $72.10$ & $67.09$ & $60.31$ & $55.62$ & $54.60$ & $50.42$ \\
\cellcolor{lightPurple}\textbf{\ourmethodCC}\textcolor{gray}{\scriptsize{ [Ours, NeurIPS'25]}} & 
\cellcolor{lightPurple}$\underline{60.11}$ & \cellcolor{lightPurple}$\underline{57.40}$ & \cellcolor{lightPurple}$\underline{53.34}$ & \cellcolor{lightPurple}$\underline{50.95}$ & \cellcolor{lightPurple}$\underline{78.94}$ & \cellcolor{lightPurple}$\underline{74.80}$ & \cellcolor{lightPurple}$\underline{73.91}$ & \cellcolor{lightPurple}$\underline{70.14}$ & \cellcolor{lightPurple}$\underline{62.05}$ & \cellcolor{lightPurple}$\underline{58.31}$ & \cellcolor{lightPurple}$\underline{56.05}$ & \cellcolor{lightPurple}$\underline{52.80}$ \\
\cellcolor{lightPurple}\textbf{MAIL++}\textcolor{gray}{\scriptsize{ [Ours]}} & 
\cellcolor{lightPurple}$\textbf{61.38}$ & \cellcolor{lightPurple}$\textbf{58.76}$ & \cellcolor{lightPurple}$\textbf{54.69}$ & \cellcolor{lightPurple}$\textbf{52.53}$ & \cellcolor{lightPurple}$\textbf{79.47}$ & \cellcolor{lightPurple}$\textbf{75.48}$ & \cellcolor{lightPurple}$\textbf{74.33}$ & \cellcolor{lightPurple}$\textbf{70.81}$ & \cellcolor{lightPurple}$\textbf{63.23}$ & \cellcolor{lightPurple}$\textbf{59.63}$ & \cellcolor{lightPurple}$\textbf{57.13}$ & \cellcolor{lightPurple}$\textbf{54.13}$ \\
\bottomrule[1.1pt]
\end{tabular}
}
\label{tab: UCDR results}
\vspace{-0.5\baselineskip}
\end{table*}

\vspace{-0.2\baselineskip}
\subsection{Few-Shot Classification Tasks and Datasets}

\subsubsection{\textbf{Base-to-Novel Evaluation}}
Under this evaluation protocol, the categories of each dataset are evenly divided into \emph{base} and \emph{novel} sets. Model adaptation is performed solely on the base classes, and performance is evaluated on both base and novel categories, enabling an assessment of both transfer to seen classes and the preservation of the inherent generalization and zero-shot discrimination capabilities of pretrained vision–language models on unseen classes.
We conduct experiments under this setting on eleven representative image classification benchmarks, including ImageNet~\cite{imagenet}, Caltech101~\cite{caltech101}, OxfordPets~\cite{oxford_pets}, StanfordCars~\cite{stanford_cars}, Flowers102~\cite{flowers102}, Food101~\cite{food101}, FGVCAircraft~\cite{fgvc_aircraft}, SUN397~\cite{sun397}, UCF101~\cite{ucf101}, DTD~\cite{dtd}, and EuroSAT~\cite{eurosat}. More details of the datasets can be found in the \textbf{Supplementary Material}.

\subsubsection{\textbf{Cross-Dataset Evaluation}} This evaluation protocol measures the model’s ability to generalize across previously unseen datasets. Following the CoCoOp protocol~\cite{CoCoOp}, the model is trained on all 1,000 ImageNet categories with only a few samples per class and is then directly evaluated on the same target datasets used in the Base-to-Novel setting, without any further adaptation, to assess cross-dataset transferability.

\subsubsection{\textbf{Domain Generalization Evaluation}} Analogous to the cross-dataset setting, the model is trained on ImageNet and evaluated on four domain-shifted variants—ImageNetV2~\cite{imagenetv2}, ImageNet-Sketch~\cite{imagenet_sketch}, ImageNet-A~\cite{imagenet_a}, and ImageNet-R~\cite{imagenet_r}, enabling a complementary assessment. 

\vspace{-0.5\baselineskip}
\subsection{Few-Shot Retrieval Tasks}

\subsubsection{\textbf{FS-UCDR}}
The few-shot universal cross-domain retrieval (FS-UCDR) protocol was first introduced in our prior work~\cite{MAIL}. Under this setting, the shared label space is instantiated with only a few samples per class from each source domain, and the model is trained on sparsely supervised multi-domain data and subsequently evaluated on cross-domain retrieval setting, where the gallery is drawn from the real domain and query samples originate from previously unseen domains and unseen classes. By default, the gallery classes are identical to those of the query set, which we denote as the $\rm Unseen Gallery$. In more realistic scenarios, the gallery may further include additional classes—such as those observed during training—referred to as the $\rm Mixed Gallery$.
 

\subsubsection{\textbf{FS-U$^{\mathbf{D}}$CDR}}
Under the few-shot universal domain cross-domain retrieval (FS-U$^{D}$CDR), the queried categories have been encountered during training, yet the query samples arise from novel, unseen domains. This protocol evaluates robustness to domain shift under controlled semantic novelty.


\subsubsection{\textbf{FS-U$^{\mathbf{C}}$CDR}}
Under the few-shot universal class cross-domain retrieval (FS-U$^{C}$CDR), the model is exposed to the domains during training but must retrieve semantically aligned samples for classes absent from the training set. This evaluation stresses the preservation of semantic generalization.

We conduct experiments on three datasets: DomainNet~\cite{DomainNet}, Sketchy~\cite{Sketchy-1, Sketchy-2}, and TU-Berlin~\cite{Berlin-1, Berlin-2}. DomainNet is used for the FS-UCDR and FS-U$^{D}$CDR evaluations, while Sketchy and TU-Berlin are adopted for FS-U$^{C}$CDR. More details  are provided in the \textbf{Supplementary Material}.

\subsection{Implementation Details}
\subsubsection{Implementation Details for Classification Tasks}
Following prior work~\cite{CoOp, CoCoOp, MaPLe, MMA}, we adopt CLIP with a ViT-B/16~\cite{ViT} backbone for all experiments, where the text and visual embedding dimensions are set to $d_t \! = \! 512$ and $d_v \! = \! 768$, respectively. For MAIL++, the dimension of the meta-scaling vector $a_m$ is set to 512. The configuration for $\lambda$ is provided in the \textbf{Supplementary Material}.
Hand-crafted text prompts from prior methods~\cite{CLIP, CoOp, Tip-adapter} are utilized. A 16-shot training strategy is employed, where 16 samples per class are randomly selected. 
For \textbf{Base-to-Novel Evaluation}, we use a batch size of 64 for larger datasets (ImageNet and SUN397), 3 for the smallest dataset EuroSAT, and 4 for all remaining datasets. The model is trained for 5 epochs on ImageNet and 10 epochs on the other datasets. We adopt the AdamW optimizer for all experiments, except on EuroSAT, where SGD yields better performance. The initial learning rate is set to $1.5 \times 10^{-5}$ for all datasets. The rank ($r$) of the bridge function is set to 96 for EuroSAT and 32 for the remaining datasets.
For \textbf{Cross-Dataset evaluation and Domain Generalization tasks,}  We train the model on ImageNet for 1 epoch, while all other training settings remain the same as those used for ImageNet in the Base-to-Novel evaluation. The average accuracy is reported over three independent runs with random seeds set to 0, 1, and 2. All experiments are conducted on two 4090D GPUs 

\subsubsection{Implementation Details for Retrieval Tasks}
For UCDR tasks, the ViT-B/32 backbone~\cite{ViT} is employed following prior CLIP-based studies~\cite{ProS}, with $d_t \! = \! 512$ and $d_v \! =  \!768$. In MAIL++, the bridge function rank $r$ is set to 16 and the meta-scaling vector $a_m$ dimension is set to 512.  $\lambda$ is set to 0, \ie, the regularization term is not applied. The text prompt is fixed as ``\texttt{a photo of a $<$category$>$}''. The triplet-hard loss~\cite{MAIL}, which is generally used for retrieval tasks, is also adopted.
All experiments adhere to a 2-shot protocol, providing $2 \! \times \! N_S$ samples per category, where $N_S$ is the number of source domains ($N_S\!=\!5$ for DomainNet, $N_S\! =\! 2$ for Sketchy and TU-Berlin).
Training spans 1 epoch for DomainNet and 20 epochs for Sketchy and TU-Berlin, with early stopping after 2 epochs. To accommodate multiple source domains in FS-UCDR and the utilization of triple-hard loss~\cite{MAIL}, each batch is organized as $B = N_S \! \times \! C_b \! \times \! k_b$, where $C_b \! = \! 3$ denotes the number of classes sampled per domain, and $k_b = 4$ the number of images per class; repeated samples occur if $k_b > k$, with $k$ being the shot number. Optimization employs AdamW with an initial learning rate of $2 \! \times \! 10^{-4}$ and a cosine decay schedule. All experiments are conducted on a single 4090D GPU with mixed-precision training.

\subsection{Comparision Results under Classification Tasks}

\subsubsection{Base-to-Novel Evaluation}
We evaluate MAIL and MAIL++ against a comprehensive set of competitive baselines, including the zero-shot CLIP model and representative state-of-the-art prompt-learning approaches, namely CoOp~\cite{CoOp}, CoCoOp~\cite{CoCoOp}, KgCoOp~\cite{KgCoOp}, MaPLe~\cite{MaPLe}, PromptSRC~\cite{PromptSRC}, TCP~\cite{TCP}, DeKg~\cite{DeKg}, BIP-D~\cite{BIP-D}, and FCPrompt-G~\cite{FCPrompt-G}, as well as the multimodal adapter-based method MMA~\cite{MMA}. To ensure a fair and controlled comparison, we deliberately exclude methods that rely on large language models (LLMs) to introduce strong prompt priors~\cite{MMA++, MaPLe++} or employ distillation strategies that exploit the full unlabeled target dataset~\cite{PromptKDcvpr}, as these settings introduce additional supervision beyond the standard few-shot protocol. \textbf{Tab.~\ref{tab: base_to_novel}} provides detailed results for Base and Novel classes across 11 datasets, along with the balanced harmonic mean (HM) of their accuracies, yielding several key insights: \ding{182} \textit{MAIL++ achieves a new SOTA performance.}  MAIL and MAIL++ consistently achieve superior or highly competitive HM performance across most datasets, indicating a better balance between base and novel classes. Specifically,  MAIL++ yields improvements of 1.57\%, 0.86\% on Novel and HM metrics respectively, over the previous best model MMRL. \ding{183} \textit{Strong advantages on fine-grained and low-data datasets.} MAIL and MAIL++ show particularly strong performance on fine-grained and data-sensitive datasets, such as OxfordPets, StanfordCars, FGVCAircraft, and Flowers102. On these datasets, existing prompt-based methods often exhibit large performance variance, whereas MAIL maintains stable and competitive results, highlighting its ability to capture discriminative local semantics under few-shot settings.
\ding{184} \textit{MAIL++ brings consistent gains over MAIL, especially on novel classes.} Compared to MAIL, MAIL++ yields consistent improvements in Novel accuracy and HM across nearly all benchmarks. This consistent trend highlights the critical role of bidirectional information flow between accross the modalities.


\subsubsection{Cross-Dataset Evaluation}
The quantitative results in \textbf{Tab.~\ref{tab: cross_dataset}} indicate that the proposed methods generalize well to unseen target datasets. In particular, MAIL++ achieves an average accuracy of 67.98\%, improving upon previously reported state-of-the-art results, including MMRL (67.25\%) and Dekg (66.64\%). 
Although MAIL and MAIL++ do not attain the best performance on ImageNet, the performance gap with Dekg is marginal (only 0.23\%), which is considered acceptable given the overall improvements across diverse datasets.

\subsubsection{Domain Generalization Evaluation}
As summarized in \textbf{Tab.~\ref{tab: domain_generalization}}, MAIL and MAIL++ achieve state-of-the-art performance on the majority of evaluated datasets. In particular, MAIL++ attains an average accuracy of 61.26\%, exceeding the previous best-performing method MMRL by 1.60\%. 

\begin{table*}[t]
\centering
\vspace{-0.3\baselineskip}
\caption{FS-U$^D$CDR (2-shot) evaluation results (\%) on DomainNet.}
\renewcommand\arraystretch{0.92} 
\setlength{\tabcolsep}{4.3pt}
\resizebox{0.92\linewidth}{!}{
    \begin{tabular}{l|cc|cc|cc|cc|cc|cc}
    \toprule[1.1pt]

        \textbf{\multirow{2}{*}[-0.5ex]{Methods}}   &  \multicolumn{2}{c|}{\textit{\textbf{Sketch}}} & \multicolumn{2}{c|}{\textit{\textbf{Quickdraw}}} & \multicolumn{2}{c|}{\textit{\textbf{Painting}}} & \multicolumn{2}{c|}{\textit{\textbf{Infograph}}} &\multicolumn{2}{c|}{\textit{\textbf{Clipart}}} & \multicolumn{2}{c}{\textit{\textbf{Average}}}  \\
       \cmidrule(lr){2-3} \cmidrule(lr){4-5} \cmidrule(lr){6-7} \cmidrule(lr){8-9} \cmidrule(lr){10-11} \cmidrule(lr){12-13} 
        & mAP$_{200}$ & Prec$_{200}$ & mAP$_{200}$ & Prec$_{200}$ & mAP$_{200}$ & Prec$_{200}$ & mAP$_{200}$ & Prec$_{200}$ & mAP$_{200}$ & Prec$_{200}$ & mAP$_{200}$ & Prec$_{200}$ \\
        \cmidrule(lr){1-1} \cmidrule(lr){2-3} \cmidrule(lr){4-5} \cmidrule(lr){6-7} \cmidrule(lr){8-9} \cmidrule(lr){10-11} \cmidrule(lr){12-13} 
        ProS*\textcolor{gray}{\scriptsize{ [CVPR'24]}}
        & $73.85$ & $49.11 $ & $ 28.89 $ & $ 11.86 $ & $ 72.27 $ & $ 46.15 $ & $60.56$ & $39.62$ & $81.05$ & $52.98$ & $63.32$ & $39.94$ \\
        \cmidrule(lr){1-1} \cmidrule(lr){2-3} \cmidrule(lr){4-5} \cmidrule(lr){6-7} \cmidrule(lr){8-9} \cmidrule(lr){10-11} \cmidrule(lr){12-13} 
        CLIP\textcolor{gray}{\scriptsize{ [ICML'21]}} 
        & $47.60$ & $28.71$ & $8.67$  & $4.50$  & $55.69$ & $31.70$ & $47.56$ & $29.36$ & $55.81$ & $31.10$ & $43.07$ & $25.07$\\
        BitFit\textcolor{gray}{\scriptsize{ [ACL'22]}} 
        & $70.20$ & $45.00$ & $22.75$ & $9.52$  & $68.54$ & $41.66$ & $59.77$ & $38.51$ & $75.17$ & $47.09$ & $59.29$ & $36.36$\\
        VPT-D\textcolor{gray}{\scriptsize{ [ECCV'22]}} 
        & $65.63$ & $42.96$ & $22.33$ & $10.01$ & $66.66$ & $41.13$ & $57.63$ & $37.20$ & $73.64$ & $46.46$ & $57.18$ & $35.55$  \\
        AFormer\textcolor{gray}{\scriptsize{ [NeurIPS'22]}} 
        & $66.28$ & $39.91$ & $20.66$ & $7.88$  & $64.77$ & $37.21$ & $59.55$ & $36.70$ & $70.51$ & $52.16$ & $56.35$ & $34.77$  \\
        IVLP\textcolor{gray}{\scriptsize{ [CVPR'23]}}  
        & $64.54$ & $42.17$ & $21.31$ & $9.51$  & $66.55$ & $40.58$ & $59.26$ & $37.64$ & $72.31$ & $44.80$ & $56.79$ & $34.94$ \\
        IVLA\textcolor{gray}{\scriptsize{ [CVPR'24]}} 
        & $66.58$ & $40.18$ & $21.44$ & $8.20$  & $64.84$ & $37.31$ & $58.42$ & $37.58$ & $70.66$ & $42.29$ & $56.39$ & $33.11$ \\
        MaPLe\textcolor{gray}{\scriptsize{ [CVPR'23]}} 
        & $71.18$ & $46.65$ & $26.48$ & $11.28$ & $70.72$ & $44.64$ & $60.24$ & $39.40$ & $77.47$ & $49.15$ & $61.22$ & $38.22$ \\
        MMA\textcolor{gray}{\scriptsize{ [CVPR'24]}}  
        & $71.14$ & $45.54$ & $24.03$ & $10.17$ & $68.66$ & $41.53$ & $59.52$ & $36.74$ & $75.92$ & $46.38$ & $59.86$ & $36.07$\\
        \cellcolor{lightPurple}\textbf{\ourmethodCC}\textcolor{gray}{\scriptsize{ [Ours, NeurIPS'25]}} &
        \cellcolor{lightPurple}$\underline{73.61}$ & \cellcolor{lightPurple}$\underline{49.20}$ & \cellcolor{lightPurple}$\underline{26.91}$ & \cellcolor{lightPurple}$\underline{11.60}$ & \cellcolor{lightPurple}$\underline{72.53}$ & \cellcolor{lightPurple}$\underline{45.93}$ & \cellcolor{lightPurple}$\underline{62.69}$ & \cellcolor{lightPurple}$\underline{41.35}$ & \cellcolor{lightPurple}$\underline{79.81}$ & \cellcolor{lightPurple}$\underline{52.21}$ & \cellcolor{lightPurple}$\underline{63.11}$ & \cellcolor{lightPurple}$\underline{40.06}$ \\
        \cellcolor{lightPurple}\textbf{MAIL++}\textcolor{gray}{\scriptsize{ [Ours]}}  & \cellcolor{lightPurple}$\textbf{74.96}$ & \cellcolor{lightPurple}$\textbf{49.89}$ & \cellcolor{lightPurple}$\textbf{28.10}$ & \cellcolor{lightPurple}$\textbf{11.64}$ & \cellcolor{lightPurple}$\textbf{73.08}$ & \cellcolor{lightPurple}$\textbf{46.29}$ & \cellcolor{lightPurple}$\textbf{63.81}$ & \cellcolor{lightPurple}$\textbf{41.68}$ & \cellcolor{lightPurple}$\textbf{80.25}$ & \cellcolor{lightPurple}$\textbf{52.44}$  & \cellcolor{lightPurple}$\textbf{64.04}$ & \cellcolor{lightPurple}$\textbf{40.39}$ \\
    \bottomrule[1.1pt]
    \end{tabular}
}
\label{tab: UDCDR results}
\vspace{-0.5\baselineskip}
\end{table*}

\begin{table}[t]
\centering
\vspace{-0.3\baselineskip}
\caption{ FS-U$^C$CDR (2-shot) evaluation results (\%) on Sketchy and TU-Berlin. }
\renewcommand\arraystretch{0.92} 
\setlength{\tabcolsep}{6.5pt}
\resizebox{0.9\columnwidth}{!}{
    \begin{tabular}{l|cc|cc}
        \toprule[1.1pt]
        \multirow{2}{*}[-0.35ex]{\textbf{Methods}}   
        & \multicolumn{2}{c|}{\textbf{Sketchy}} & \multicolumn{2}{c}{\textbf{TU-Berlin}} \\
        \cmidrule(lr){2-3} \cmidrule(lr){4-5}  
        & mAP$_{200}$ & Prec$_{200}$ & mAP$_{all}$ & Prec$_{100}$ \\
        \cmidrule(lr){1-1} \cmidrule(lr){2-3} \cmidrule(lr){4-5} 
        ProS*\textcolor{gray}{\scriptsize{ [CVPR'24]}}         & $69.91$ & $65.45$ & $66.75$ & $74.42$\\
        \cmidrule(lr){1-1} \cmidrule(lr){2-3} \cmidrule(lr){4-5} 
        CLIP\textcolor{gray}{\scriptsize{ [ICML'21]}}         & $35.82$ & $33.08$ & $31.45$ & $46.12$\\
        BitFit\textcolor{gray}{\scriptsize{ [ACL'22]}}        & $67.71$ & $64.01$ & $65.51$ & $73.68$\\
        VPT-D\textcolor{gray}{\scriptsize{ [ECCV'22]}}        & $65.19$ & $61.16$ & $62.12$ & $70.89$\\
        AFormer\textcolor{gray}{\scriptsize{ [NeurIPS'21]}}   & $56.87$ & $52.31$ & $58.95$ & $70.14$\\
        IVLP\textcolor{gray}{\scriptsize{ [CVPR'23]}}         & $60.27$ & $55.75$ & $59.13$ & $68.58$\\
        IVLA\textcolor{gray}{\scriptsize{ [CVPR'24]}}         & $56.77$ & $52.71$ & $59.13$ & $70.42$\\
        MaPLe\textcolor{gray}{\scriptsize{ [CVPR'23]}}        & $71.86$ & $68.14$ & $65.90$ & $73.73$\\
        MMA\textcolor{gray}{\scriptsize{ [CVPR'24]}}          & $61.59$ & $57.14$ & $63.70$ & $72.69$\\
        \cellcolor{lightPurple}\textbf{MAIL}\textcolor{gray}{\scriptsize{ [Ours, NeurIPS'25]}}  &
        \cellcolor{lightPurple}\underline{$73.46$} &
        \cellcolor{lightPurple}\underline{$69.73$} &
        \cellcolor{lightPurple}\underline{$67.97$} & 
        \cellcolor{lightPurple}\underline{$75.10$} \\
        \cellcolor{lightPurple}\textbf{MAIL++}\textcolor{gray}{\scriptsize{ [Ours]}}  &
        \cellcolor{lightPurple}\textbf{74.07} &
        \cellcolor{lightPurple}\textbf{70.15} &
        \cellcolor{lightPurple}\textbf{68.78} & 
        \cellcolor{lightPurple}\textbf{75.56} \\
        \bottomrule[1.1pt]
    \end{tabular}
}
\label{tab: UCCDR results}
\vspace{-0.5\baselineskip}
\end{table}

\subsection{Comparison Results under Retrieval Tasks}
\subsubsection{Comparison Results under FS-UCDR and FS-U$^D$CDR} 
 We primarily compare our MAIL and MAIL++ with the following categories: \ding{182} adapter-based methods, including AdaptFormer (vision-only adapter)~\cite{Adaptformer}, IVLA (modality-independent adapter)~\cite{MMA}, and MMA~\cite{MMA}; \ding{183} prompt-based methods, including VPT-D (vision-only prompt)~\cite{VPT}, IVLP (modality-independent prompt)~\cite{MaPLe}, and MaPLe~\cite{MaPLe}; \ding{184} other methods, including the zero-shot CLIP~\cite{CLIP}, BitFit~\cite{BitFit}, and the  full-shot UCDR method, ProS~\cite{ProS}. The results are summarized in \textbf{Tab.~\ref{tab: UCDR results}} and \textbf{Tab.~\ref{tab: UDCDR results}}.
We identify several key observations:
\ding{182} \textit{Our methods consistently outperforms existing baselines.} Under both settings, MAIL++ achieves clear performance advantages over ProS, despite requiring only about 1/140 of the training data used by ProS, indicating markedly superior data efficiency and generalization capability in data-scarce UCDR scenarios.
\ding{183} \textit{Modality-coupled methods consistently outperform modality independent methods.} For instance, under FS-UCDR's $\rm UnseenGallery$ scenario, MaPLe and MMA achieve average mAP$_{200}$ improvements of 4.44\% and 3.22\% over IVLP and IVLA, respectively. This emphasizes the importance of collaboration and information sharing between modalities in low-data settings.
\ding{184} \textit{The vision-only methods perform similarly to, or even slightly outperform, the modality-independent methods.}
As seen in the table, AdaptFormer~\cite{Adaptformer} achieves results comparable to IVLA, while VPT-Deep achieves an average mAP improvement of 0.5\%-1.6\% over IVLP. Therefore, we conclude that the benefit of simply fine-tuning the text side for retrieval is limited. 

\subsubsection{Comparison Results under FS-U$^C$CDR} \textbf{Tab.~\ref{tab: UCCDR results}} compares the FS-U$^C$CDR performance of our MAIL and MAIL++ with other baselines. The results demonstrate that MAIL++ consistently achieves the best performance among all methods, indicating its effectiveness in enhancing CLIP’s capability to handle semantic shifts under limited-data scenarios.

\subsection{Ablation Studies}

In this section, we systematically investigate the effectiveness of the proposed modules in MAIL++ through comprehensive ablation studies. Unless otherwise specified, all experiments are conducted under the Base-to-Novel classification setting, and the results are averaged over all eleven datasets.


\begin{table}[t]
    \centering
    \caption{Ablation on different information flow directions.}
    \vspace{-0.3\baselineskip}
    \renewcommand\arraystretch{0.92} 
    \setlength{\tabcolsep}{10pt}
    \label{tab: mail++ variant}
    \resizebox{0.85\columnwidth}{!}{
    \begin{tabular}{lc|ccc}
        \toprule[1.1pt]
        AL Position &  Direction & Base &  Novel & HM \\
        \cmidrule(lr){1-2} \cmidrule(lr){3-5} 
        Only Text     & None                   & $78.88$ & $76.18$ & $77.51$  \\
        Only Image    & None                   & $77.12$ & $76.47$ & $76.79$  \\
        Image \& Text & None                   & $80.20$ & $77.30$ & $78.72$  \\
        Image \& Text & T $\Rightarrow$ I      & $84.61$ & $77.64$ & $80.98$ \\
        Image \& Text & I $\Rightarrow$ T      & $84.67$ & $77.22$ & $80.77$ \\   
        \rowcolor{lightPurple}
        Image \& Text & T $\Leftrightarrow$ I  & \textbf{85.69} & \textbf{78.73} & \textbf{82.06} \\  
        \bottomrule[1.1pt]
    \end{tabular}
    }
\vspace{-1.0\baselineskip}
\end{table}

\begin{figure}[t]
\begin{center}
\centerline{\includegraphics[width=.9\columnwidth]{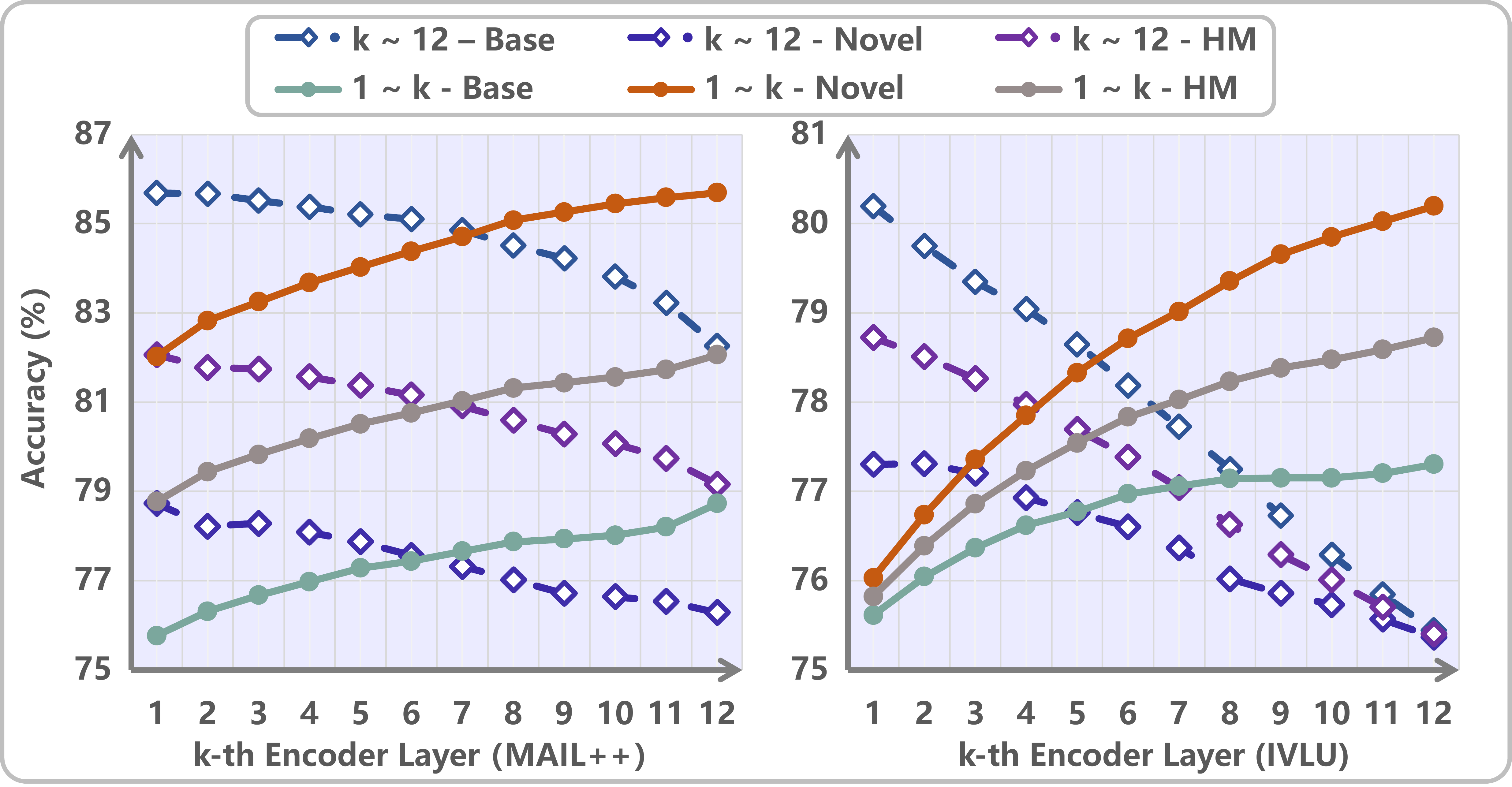}}
\vspace{-0.3\baselineskip}
\caption{Ablation studies on insertion strategies across Transformer blocks. }
\label{fig: ablation_depth}
\end{center}
\vspace{-1.2\baselineskip}
\end{figure}

\subsubsection{\textbf{Variants on Different Information Flow Directions}} We first evaluate the performance of different possible inflomation flow direction design choices in MAIL++. As shown in \textbf{Tab.~\ref{tab: mail++ variant}}, the agent layer (AL) setting consistently outperforms unimodal variants, highlighting the necessity of integrating agent layers into both encoders. Enabling directional cross-modal information flow further improves performance, while the bidirectional design (T $\Leftrightarrow$ I) achieves the best results on both Base and Novel sets, leading to the highest HM. These results indicate that mutual information exchange between modalities is critical for achieving robust performance.

\subsubsection{\textbf{Variants of Adding AL and MAIL++ Across the Transformer Blocks}} As shown in \textbf{Fig.~\ref{fig: ablation_depth}}, we systematically vary the depth-wise deployment of both bridged agent layers (MAIL++) and naive agent layers (IVLU) across the Transformer blocks. Specifically, agent layers are inserted either progressively from the first block up to the $k$-th block ($k = 1, 2, \ldots, 12$), or from the $k$-th block to the final block. As illustrated, the performance peaks when ALs and MAIL++ are applied to all transformer blocks.


\begin{figure*}[t]
\begin{center}
\centerline{\includegraphics[width=0.9\textwidth]{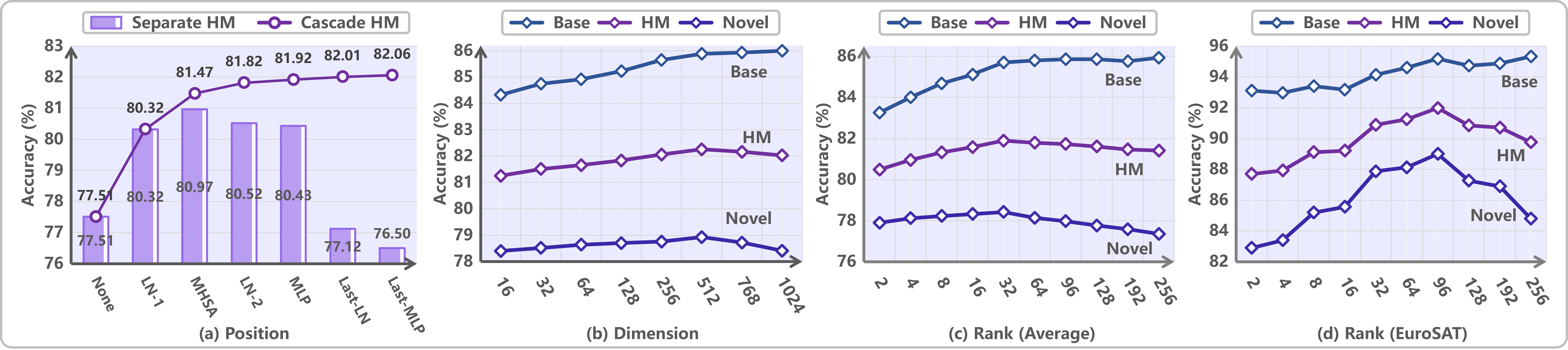}}
\vspace{-0.3\baselineskip}
\caption{Ablation studies on insertion strategies within Transformer blocks, the dimension of the meta-scaling vector, and the rank of the bridge function.}
\label{fig: position&dimension}
\end{center}
\vspace{-1.0\baselineskip}
\end{figure*}


\begin{figure}[t]
\begin{center}
\centerline{\includegraphics[width=0.9\columnwidth]{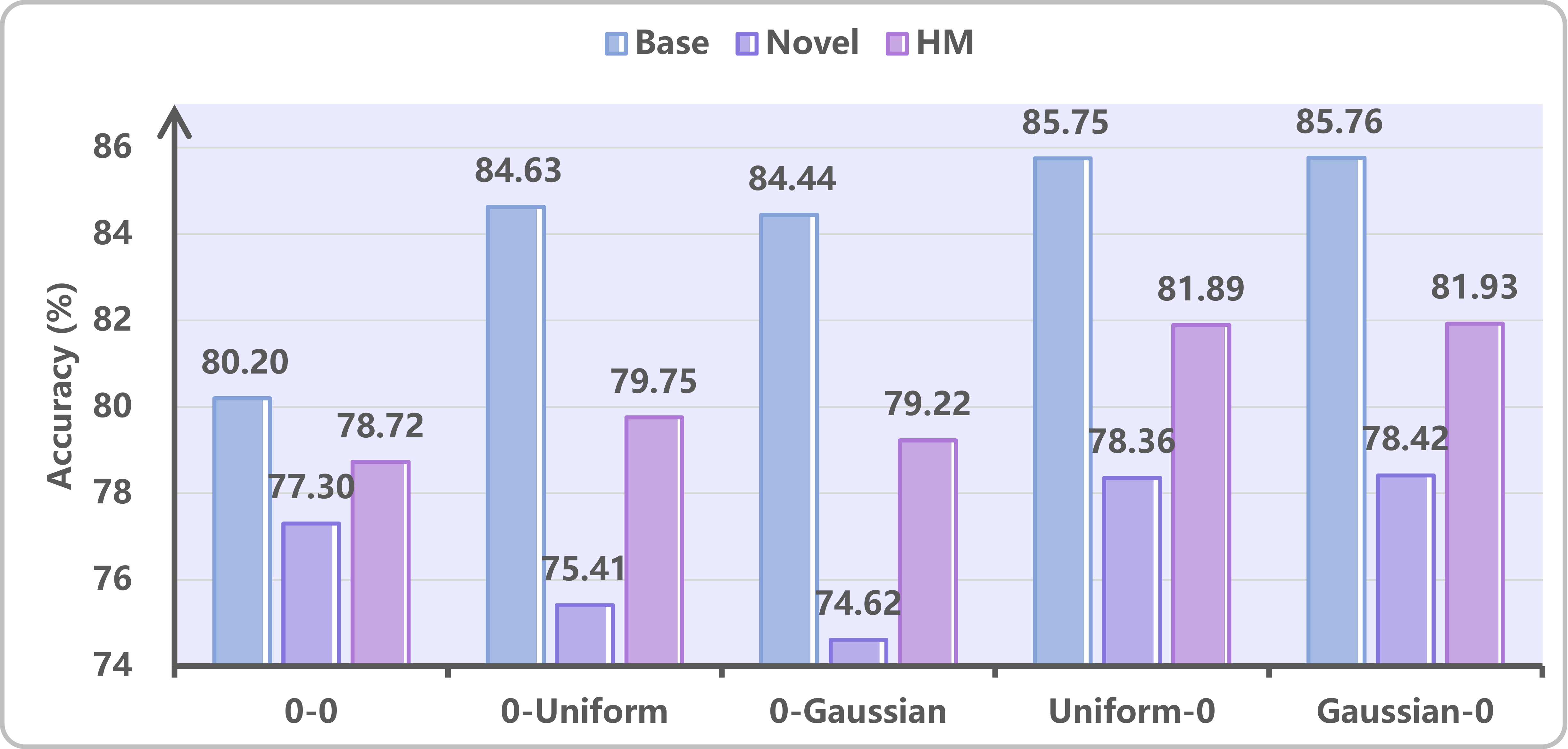}}
\vspace{-0.3\baselineskip}
\caption{Ablation studies on the initialization methods of the bridge function.}
\label{fig: init}
\end{center}
\vspace{-1.5\baselineskip}
\end{figure}

\subsubsection{\textbf{Variants of Adding MAIL++ Within the Transformer Blocks}} As shown in \textbf{Fig.~\ref{fig: position&dimension}} (a), the bar plot reports the HM accuracy obtained by inserting  MAIL++ module at an individual position within the Transformer blocks, while the curve illustrates the results of cumulative insertion from shallow to deep layers. When deployed in isolation, MAIL++ achieves the most pronounced improvement when inserted at the MHSA layer. Moreover, the cumulative strategy leads to a monotonic increase in HM performance, indicating that progressively enriching cross-modal interactions across layers provides complementary benefits.

\subsubsection{\textbf{Dimension of the Meta Scaling Vector}}
As illustrated in \textbf{Fig.~\ref{fig: position&dimension}} (b), the Base accuracy exhibits a monotonic improvement as the dimensionality increases, whereas the Novel accuracy peaks at a moderate dimension and deteriorates when the dimension becomes larger. As a result, the HM score achieves its maximum at a dimension of 512 on average, reflecting the optimal balance between preserving base-class knowledge and generalizing to novel classes. 

\subsubsection{\textbf{Rank of the Bridge Function}}
As shown in \textbf{Fig.~\ref{fig: position&dimension}} (c,d), Base accuracy consistently increases with larger ranks, whereas Novel accuracy reaches its peak at a moderate rank and degrades thereafter. Consequently, the HM score attains its optimum at rank 32 on Average and rank 96 on EuroSAT, achieving the best trade-off between base-class retention and novel-class generalization. 


\subsubsection{\textbf{Initialization Method of the Bridge Function}} 
\textbf{Fig.~\ref{fig: init}} presents an ablation study on the initialization strategies of the bridge function. The notation “A–B” denotes the initialization schemes applied to the first and second projection matrices of the bridge function, respectively. We consider three initialization strategies: all-zero initialization, uniform distribution, and Gaussian distribution. As shown, Uniform–0 and Gaussian–0 consistently achieves higher HM performance.
     \subsection{Discussion}

\begin{table}[t]
    \centering
    \caption{Ablation studies on bridge function applied to the shifting vector.}
    \vspace{-0.5\baselineskip}
    \renewcommand\arraystretch{0.92} 
    \setlength{\tabcolsep}{8pt}
    \label{tab: shift_vector_bridge_function}
    \resizebox{0.9\columnwidth}{!}{
    \begin{tabular}{lc|ccc}
        \toprule[1.05pt]
        Setup & Params. & Base &  Novel & HM \\
        \cmidrule(lr){1-2} \cmidrule(lr){3-5} 
        $w$ Bridge Function      & $7.54$M     & $85.65$ & $77.56$ & $81.41$  \\
        \rowcolor{lightPurple}
        $w/o$ Bridge Function    & $3.83$M    & \textbf{85.69} & \textbf{78.73} & \textbf{82.06} \\  
        \bottomrule[1.05pt]
    \end{tabular}
    }
\vspace{-1.0\baselineskip}
\end{table}

\begin{table}[t]
    \centering
    \caption{All methods are evaluated on a single NVIDIA RTX 4090D GPU on the ImageNet dataset. Each model is initially reproduced from publicly available implementations with the default settings reported in the corresponding papers (MaPLe: batch size 4; MMA: 128; MMRL: 32). For fair comparison, we re-implement all methods with a unified batch size of 64. During inference, FPS and test time are measured using a batch size of 100.}
    \vspace{-0.3\baselineskip}
    \renewcommand\arraystretch{0.92} 
    \setlength{\tabcolsep}{1.0pt}
    \label{tab: computationcost}
    \resizebox{0.9\columnwidth}{!}{
    \begin{tabular}{lc|cccc|r}
        \toprule[1.05pt]
        Methods & Params. & Train Time & Train Time  & Test Time & FPS & HM \\
                &  (M)      & (ms / image) & (s / all)  & (s / all) & - & (\%) \\   
        \cmidrule(lr){1-7} 
        \multicolumn{7}{c}{Original Training Settings.}\\
        \cmidrule(lr){1-7} 
        MaPLe (4)     & $3.55$ & $46.11$ & $1844.34$ & $40.25$ & $621.12$ & $73.47$ \\
        MMA (128)     & $0.68$ & $1.40$  & $112.19$  & $43.08$ & $580.32$ & $74.02$  \\   
        MMRL (32)     & $4.99$ & $6.59$  & $262.19$  & $42.06$ & $594.39$ & $74.45$  \\
        \cmidrule(lr){1-7} 
        \multicolumn{7}{c}{Unified Training Setting with a Batch Size of 64.}\\
        \cmidrule(lr){1-7} 
        MaPLe         & $3.55$ & $4.54$  & $181.39$  & $40.25$ & $621.12$ & $73.27$ \\
        MMA           & $0.68$ & $3.90$  & $156.05$  & $43.08$ & $580.32$ & $74.23$  \\   
        MMRL          & $4.99$ & $4.74$  & $189.55$  & $42.06$ & $594.39$ & $74.35$  \\
        
        \rowcolor{lightPurple}
        MAIL       & $0.64$  & $6.23$ & $250.63$ & $38.89$ & $642.84$ & $74.42$ \\
        \rowcolor{lightPurple}
        MAIL++     & $3.83$  & $6.37$ & $255.60$ & $38.89$ & $642.84$ & $74.86$ \\  
        \rowcolor{lightPurple}
        MAIL++\_Linear       & $32.79$ & $6.43$ & $257.33$ & $38.89$ & $642.84$ & $74.10$ \\  
        \bottomrule[1.05pt]
    \end{tabular}
    }
\vspace{-1.0\baselineskip}
\end{table}

\textit{1)} \textit{\textbf{Should the Shifting Vectors in the Agent Layers Be Connected Across Modalitie?}} We apply the bridge function \textit{only} to the modality-specific scaling vectors $a$. To validate this design choice, we conduct ablation studies by selectively introducing the bridge function to the shifting vectors $b$. As shown in \textbf{Tab.~\ref{tab: shift_vector_bridge_function}}, extending the bridge function to $b$ yields no performance improvement while doubling the number of trainable parameters. 

\textit{2) \textbf{ Computation Cost:}} We compare our methods with other Modality-Couled Methods, including MaPLe, MMA and MMRL. \textbf{Tab.~\ref{tab: computationcost}} reports a comparison of computational efficiency and performance under both the original training settings and a unified training batch size of 64 on ImageNet dataset. Under the original settings, different methods adopt substantially different batch sizes, leading to incomparable training time statistics. Under the unified setting, we identify several key observations: \ding{182} MAIL achieves the best parameter efficiency and MAIL++ attains the best HM score with only a marginal increase in training time compared to MAIL. \ding{183} Replacing the bottleneck-based bridge function in MAIL++ with a simple linear layer (MAIL++\_Linear) results in a noticeable increase in parameter count and training cost, while yielding inferior HM performance. This observation suggests that the bottleneck design of the bridge function plays a crucial role, and that naively increasing model complexity does not necessarily lead to better generalization. \ding{184} While MAIL and its variants incurs higher computational costs during training, thanks to the re-parameteriazation technique, they achieve the best inference efficiency.

\begin{figure}[t]
\begin{center}
\centerline{\includegraphics[width=0.8\columnwidth]{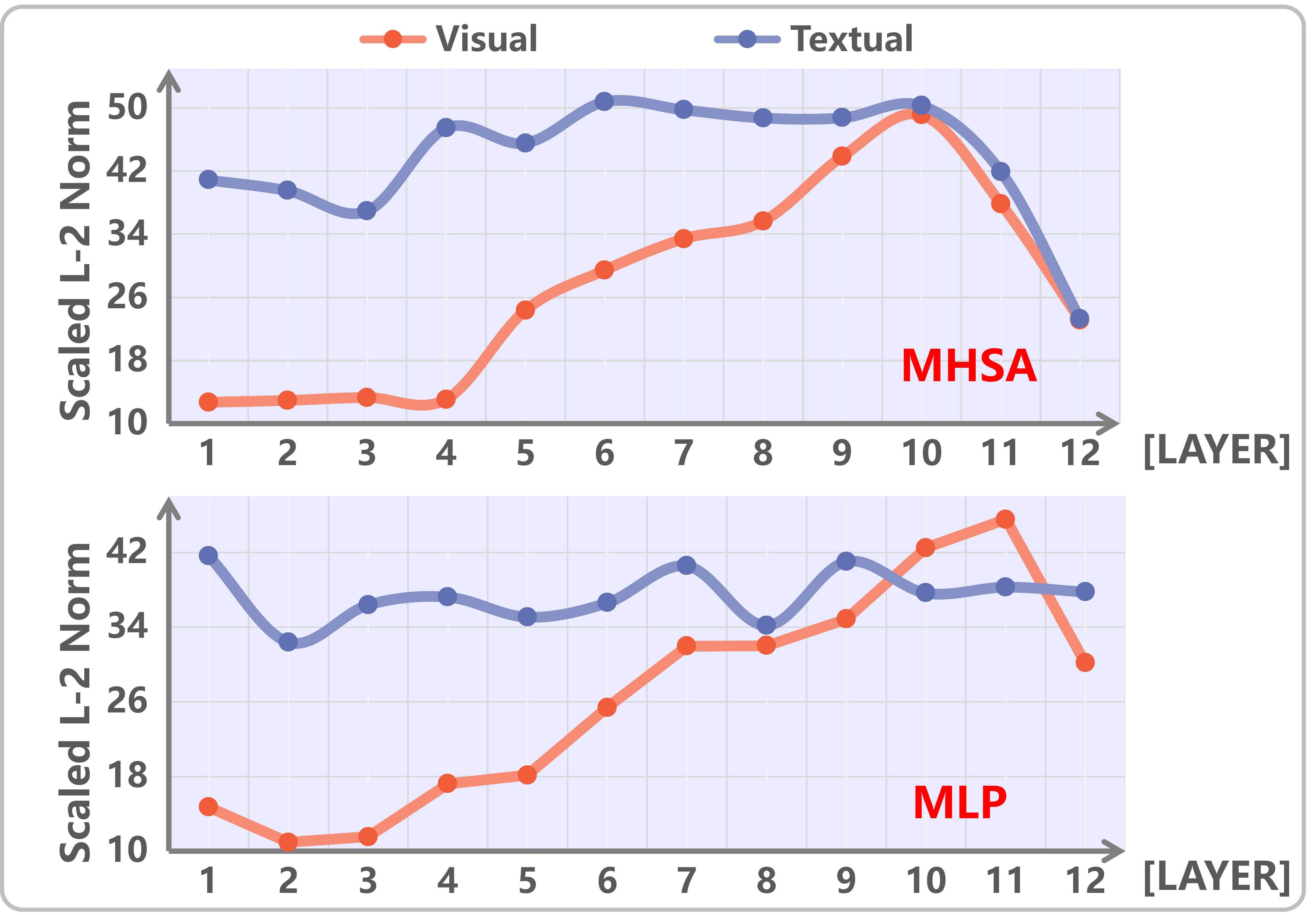}}
\vspace{-0.4\baselineskip}
\caption{Visualization of the norms of agent layers inserted after the MHSA and MLP modules across different encoder layers.}
\label{fig: norm}
\end{center}
\vspace{-1.0\baselineskip}
\end{figure}

\textit{3) \textbf{ Norm of the agent layer:}} \textbf{Fig.~\ref{fig: norm}} visualizes the scaled norms of the visual and textual modulation terms, i.e.,  $\|\frac{100}{\sqrt{d_v}} W_{up}^{v}\cdot W_{down}^{v} \cdot a_m \| $ and  $\| \frac{100}{\sqrt{d_t}} W_{up}^{t}\cdot W_{down}^{t} \cdot a_m \| $ after training in the agent layers append after  the MHSA and MLP modules across different encoder layers on the ImageNet dataset. As we can see, The model exhibits an initial predominance of textual modulation, with the contribution of the visual branch progressively increasing as more discriminative visual representations are formed. This behavior is consistent with the observation that lower layers of the visual encoder encode broadly generalizable features~\cite{MMA} and are therefore less amenable to task-specific adaptation.


\section{Conclusion}
In this work, we proposed MAIL, a modality-coupled parameter-efficient mechanism that embeds cross-modal interaction directly into the intrinsic computational modules of the CLIP backbone. By coupling localized parameter updates across the visual and textual streams, MAIL enables fine-grained and structured cross-modal alignment. We further introduced MAIL++, which removes manually imposed directional bias and facilitates balanced bidirectional interaction between modalities. Extensive experiments on few-shot classification and universal cross-domain retrieval benchmarks demonstrate that the proposed approach consistently outperforms existing PEFT methods while preserving the original inference efficiency of CLIP.







\bibliographystyle{IEEEtran}
\bibliography{reference}


 


\vspace{-9mm}

\begin{IEEEbiography}[{\raisebox{0.26\height}{\includegraphics[width=0.8in,height=1in,clip,keepaspectratio]{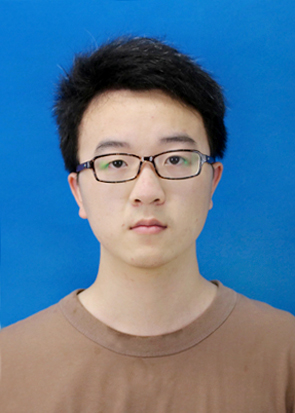}}}]{Kaixiang Chen}
 \scriptsize
 received the BSc degree from Zhejiang University of Technology, Hangzhou, China,
 in 2021, the MEng degree from the Nanjing University of Aeronautics and Astronautics, Nanjing, China, in 2024.  He is currently working toward
 the PhD degree with the School of Computer Science and Engineering, Southeast University, Nanjing, China. His research interest includes computer vision and multimedia analysis.
\end{IEEEbiography}

\vspace{-9mm}

\begin{IEEEbiography}[{\raisebox{0.26\height}{\includegraphics[width=0.8in,height=1in,clip,keepaspectratio]{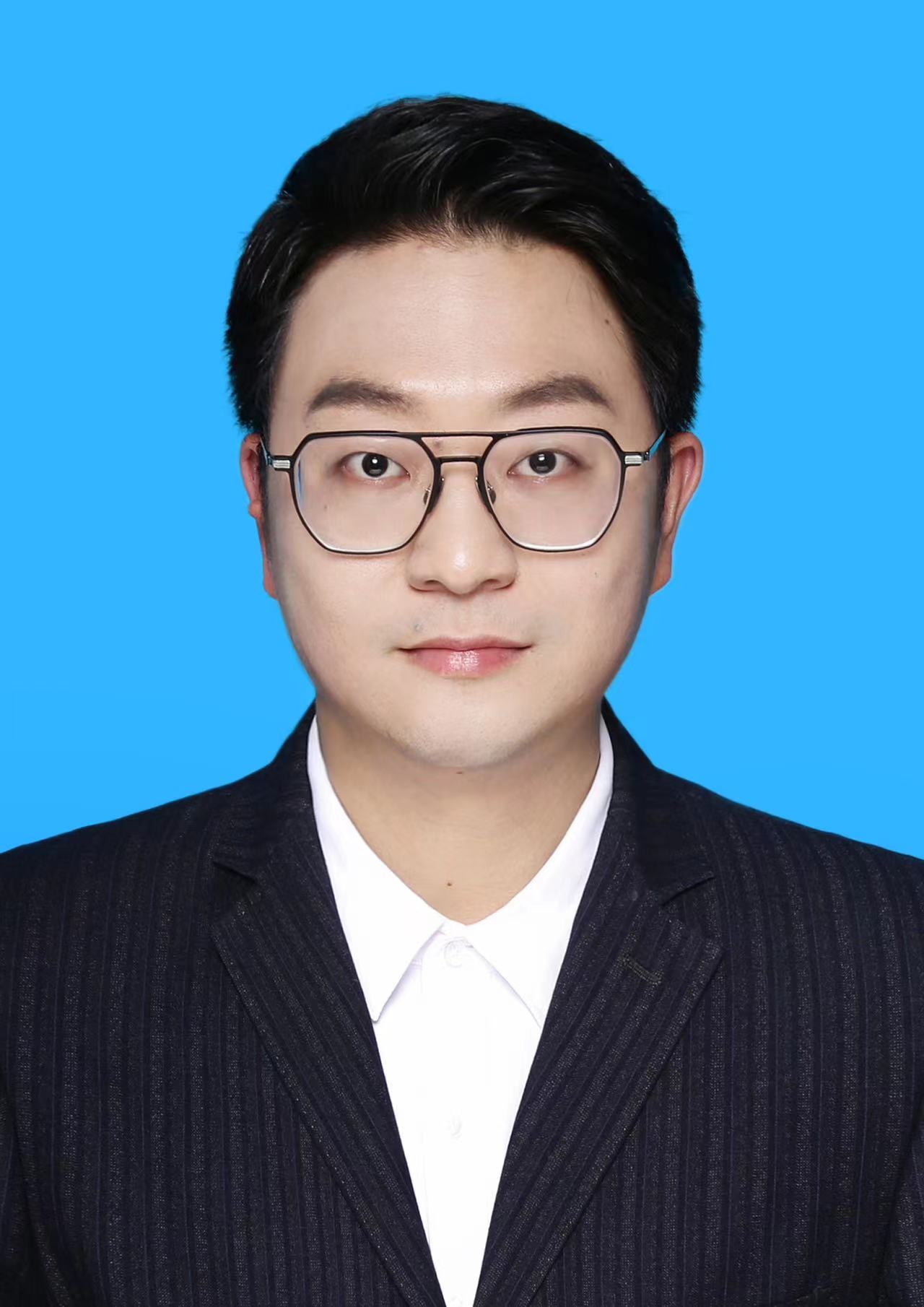}}}]{Pengfei Fang}
 \scriptsize
    is a Professor at the School of Computer Science and Engineering, Southeast University (SEU), China. Before joining SEU, he was a post-doctoral fellow at Monash University in 2022. He received the Ph.D. degree from the Australian National University and DATA61-CSIRO in 2022, and the M.E. degree from the Australian National University in 2017. His research interests include computer vision and machine learning.
\end{IEEEbiography}

\vspace{-9mm}

\begin{IEEEbiography}[{\raisebox{0.26\height}{\includegraphics[width=0.8in,height=1in,clip,keepaspectratio]{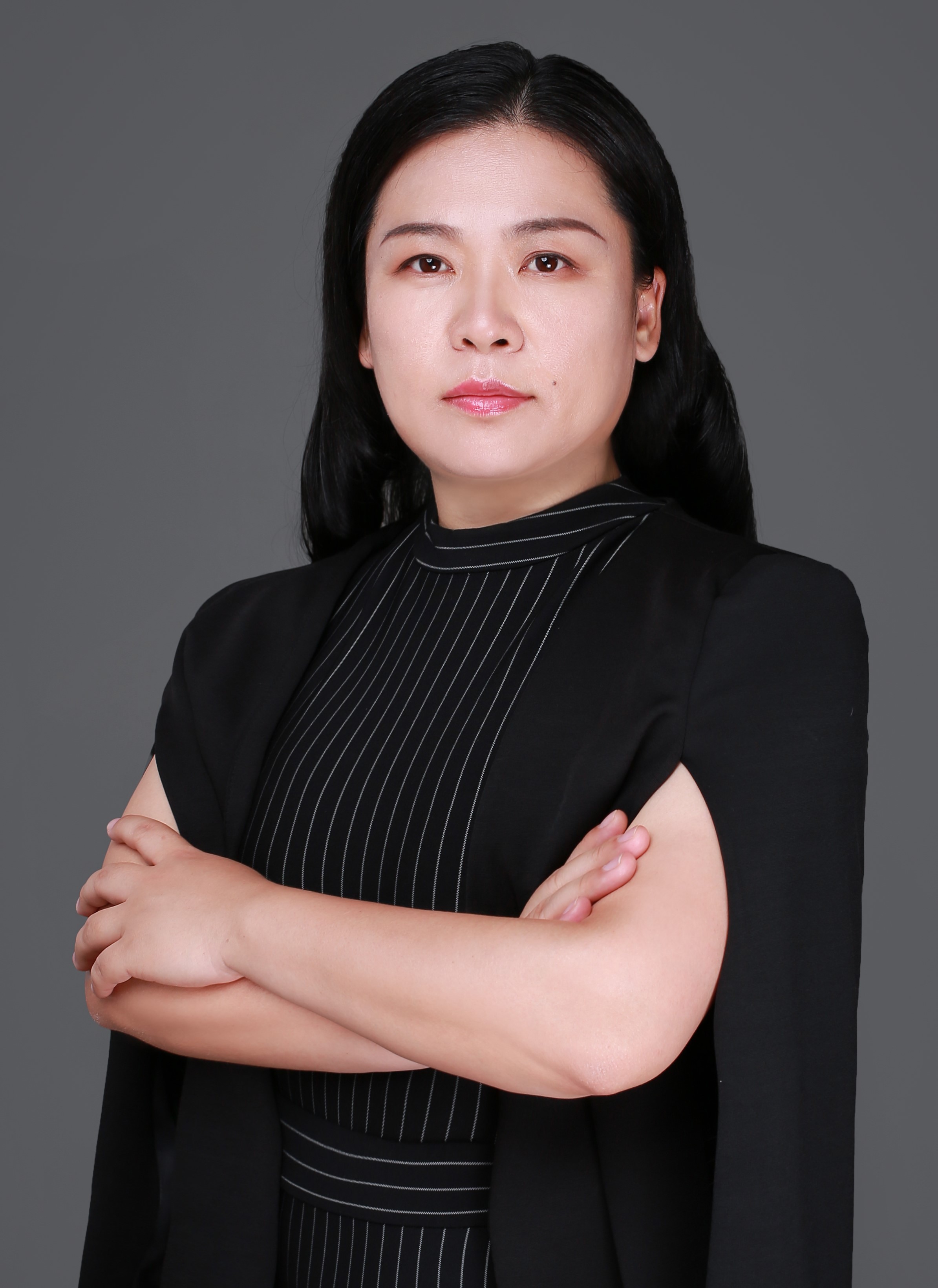}}}]{Hui Xue}(Member, IEEE)
    \scriptsize
		is currently a professor of School of Computer Science and Engineering at Southeast University, China. She received the B.Sc. degree in Mathematics from Nanjing Normal University in 2002. In 2005, she received the M.Sc. degree in Mathematics from Nanjing University of Aeronautics \& Astronautics (NUAA). And she also received the Ph.D. degree in Computer Application Technology at NUAA in 2008. Her research interests include pattern recognition and machine learning.
	\end{IEEEbiography}





\vfill

\end{document}